\documentclass[runningheads]{llncs}

\usepackage{eccv}

\usepackage{eccvabbrv}

\usepackage{graphicx}
\usepackage{booktabs}
\usepackage{multirow}
\usepackage{graphicx}
\usepackage{relsize}
\usepackage{amssymb}
\usepackage{tikz}
\usepackage{makecell}
\usepackage{marvosym}
\usepackage[accsupp]{axessibility}  % Improves PDF readability for those with disabilities.

\usepackage{hyperref}

\usepackage{orcidlink}

\begin{document}
\newcommand{\solidcircle}{%
    \begin{tikzpicture}
        \fill[black](0,0) circle (0.1cm);
    \end{tikzpicture}%
}
\newcommand{\emptycircle}{%
    \begin{tikzpicture}
        \draw[black](0,0) circle (0.1cm);
    \end{tikzpicture}%
}
\definecolor{myRed}{RGB}{219, 68, 55}
\definecolor{myGreen}{RGB}{15, 157, 88}
\definecolor{myBlue}{RGB}{66, 133, 244}
\title{CoLA: Conditional Dropout and Language-driven Robust Dual-modal Salient Object Detection} 
% \title{Unconstrained Salient and Camouflaged Object Detection} 

\titlerunning{CoLA: Dual-modal SOD}

\author{Shuang Hao\inst{1}\textsuperscript{$\star$}\orcidlink{0009-0007-4621-8485} \and
Chunlin Zhong\inst{1}\textsuperscript{$\star$}\orcidlink{0009-0006-7070-3504} \and
He Tang\inst{1}\textsuperscript{\Letter}\orcidlink{0000-0002-8454-1407}}

\authorrunning{S. Hao et al.}

\institute{School of Software Engineering, Huazhong University of Science and Technology, Wuhan, China\\
\email{\{shuanghao, clzhong, hetang\}@hust.edu.cn}
\let\thefootnote\relax\footnotetext{\textsuperscript{$\star$} Equal contribution \quad \textsuperscript{\Letter} Corresponding author}
}

\maketitle
% class-agnostic segmentation capability

% In the human visual perception mechanism, we often easily spot salient targets but may overlook camouflage objects. 
% Additionally, we introduce a new metric, \textbf{CSCS}(Camouflaged and Salient Confusion Score), to better evaluate the model's discriminative ability in distinguishing salient objects from camouflaged objects.
%这个指标比不过差距也不大，不知道为啥

\begin{abstract}
The depth/thermal information is beneficial for detecting salient object with conventional RGB images. However, in dual-modal salient object detection (SOD) model, the robustness against noisy inputs and modality missing is crucial but rarely studied. To tackle this problem, we introduce \textbf{Co}nditional Dropout and \textbf{LA}nguage-driven(\textbf{CoLA}) framework comprising two core components. 1) Language-driven Quality Assessment (LQA): Leveraging a pretrained vision-language model with a prompt learner, the LQA recalibrates image contributions without requiring additional quality annotations. This approach effectively mitigates the impact of noisy inputs. 2) Conditional Dropout (CD):  A learning method to strengthen the model's adaptability in scenarios with missing modalities, while preserving its performance under complete modalities. The CD serves as a plug-in training scheme that treats modality-missing as conditions, strengthening the overall robustness of various dual-modal SOD models. Extensive experiments demonstrate that the proposed method outperforms state-of-the-art dual-modal SOD models, under both modality-complete and modality-missing conditions. We will release source code upon acceptance.

\end{abstract}

\section{Introduction}
\label{sec:intro}

Salient object detection (SOD) is a foundational task in computer vision so as to extract the most attractive objects/regions from a scene. Benefiting from the auxiliary inputs like depth and thermal, advanced dual-modal SOD models \cite{tu2020rgbt, tu2021multi, cong2022does, wang2023thermal, zhou2023lsnet, cheng2022depth, zhang2022c} are able to detect salient objects even in complex and challenging scenes. Due to the ability of redundant elimination and reduction in computational complexity, this technique has been widely applied to enhance the performance of action recognition \cite{kong2021spatiotemporal}, object tracking \cite{zhou2021saliency,gurkan2021tdiot}, semantic segmentation \cite{chen2022saliency,lee2023saliency}, and image fusion \cite{liu2023sgfusion}.

\begin{figure}[t]
\centering
\includegraphics[width=0.9\linewidth, keepaspectratio]{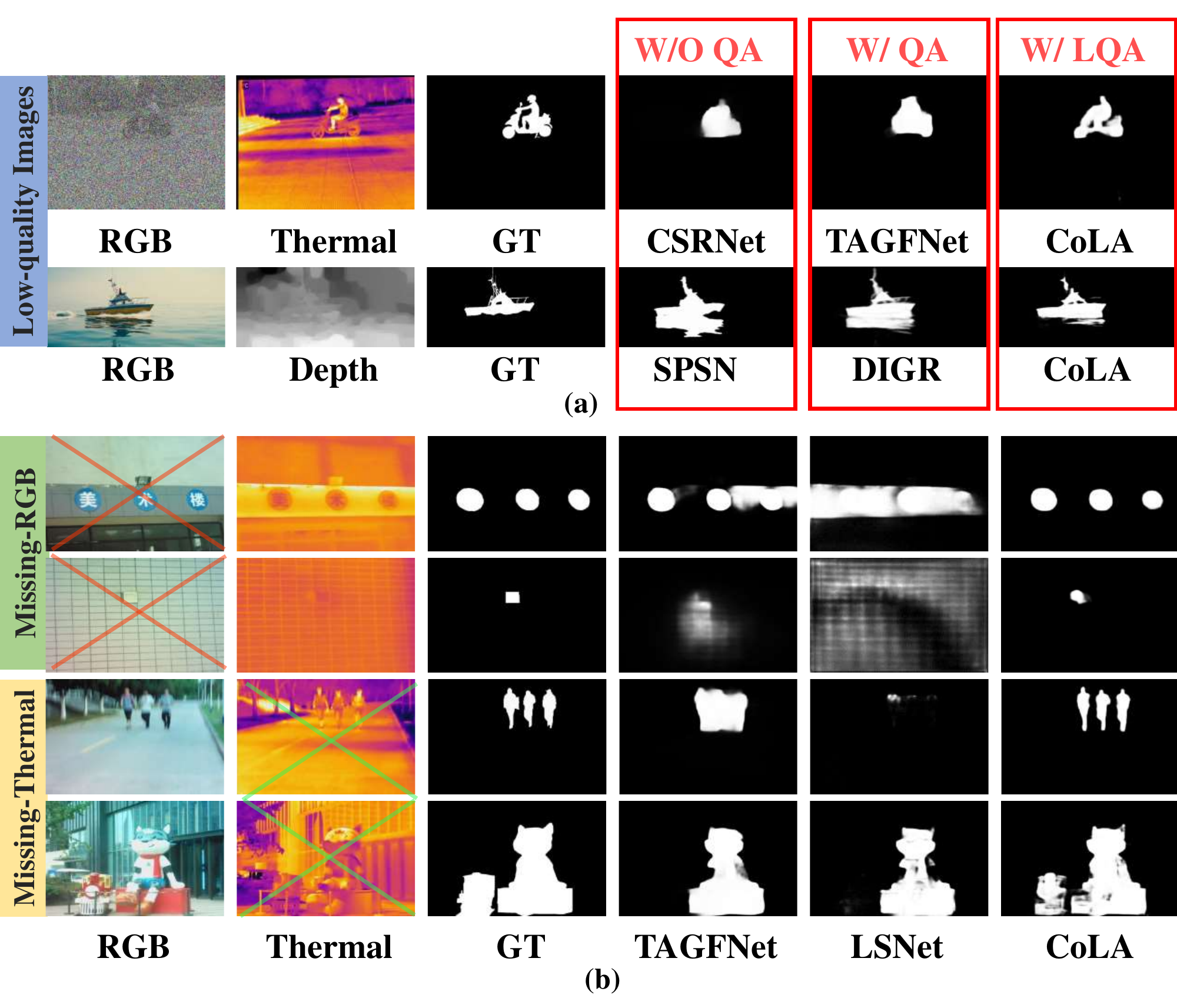}
\caption{(a): The two examples demonstrate the cases when the RGB or the Depth inputs are noisy, QA means quality assessment. (b): examples of two modality-missing conditions. The proposed CoLA produces robust results.}
\label{motivation}
\end{figure}
In general, the high accuracy of a dual-modal SOD model relies on high-quality and complete inputs. However, in real-world situations, \textit{1) the input may be noisy due to communication fault \footnote{We do not consider adversarial noise in this paper. }, 2) and one of the modalities may be missing due to device malfunction}. A SOD model trained on ideal inputs may degrade when faced with corrupted or partially missing inputs. 

As shown in Fig. \ref{motivation}(a), the noisy RGB image or depth/thermal image leads to unsatisfactory predictions in conventional methods such as CSRNet \cite{huo2021efficient} and SPSN \cite{lee2022spsn}. It is necessary to reweight the contribution of each modality rather than treating all inputs as ideal. Though quality-aware SOD models have been proposed, their quality estimation relies on either an off-the-shelf estimation network \cite{cong2022does, wang2023thermal} or pseudo-labels \cite{chen2021depth,cheng2022depth}. The parameters of the off-the-shelf estimation network \cite{cong2022does} are fixed, and pseudo-labels \cite{chen2021depth,cheng2022depth} are inaccurate; neither of them can adapt well to the target dataset such as TAGFNet \cite{wang2023thermal} and DIGR \cite{cheng2022depth}. 
Recent advances of the pre-trained vision-language model \cite{radford2021learning} have shown potential capability to assess the quality of an image \cite{wang2023exploring}. To this end, we propose a language-driven quality assessment (LQA) module to assess the quality of each input and reweight the contribution for the network.
%To overcome this limitation, we exploit the potential capability of a pre-trained vision-language model \cite{radford2021learning} and propose a language-driven quality assessment (LQA) module to assess the quality of each input and reweight the contribution for the network. 
The LQA contains a fixed prompt and a learnable prompt; only the parameters of the learnable prompt are updated during training, while the fixed prompt and vision/language encoder are frozen. Compared with other image quality assessment methods, this design not only maintains the generalization ability of the pre-trained vision-language model but also adapts to the target dataset in a parameter-efficient fine-tuning manner. As shown in the last column of Fig. \ref{motivation}(a), our predictions with the proposed LQA best fit the GT in both noisy RGB or depth.

Another limitation of existing dual-modal SOD models is that they do not consider performance degradation when the modality is partially missing. This leads to existing models overly relying on complete modal inputs, resulting in poor performance when modality-missing conditions. Furthermore, in system deployment, the phenomenon of Modality-missing occurs naturally. It should be noted that even though the RGB modality is missing from the model's input, it still exists in principle. As shown in \ref{motivation}(b), when one modality is missing, dual-modal SOD models like LSNet\cite{zhou2023lsnet} and TAGF\cite{wang2023thermal} can not recognize the salient object in the images, even though another modality image is easy to recognize the salient object. This indicates existing dual-modal SOD models lack consideration for the situation of modality-missing. Resulting in insufficient utilization of the image information of another existing modality when the modality is missing.
% Recent advancements in methods for incomplete multimodal learning have been proposed, particularly in the field of image classification \cite{wei2023mmanet} and medical image segmentation \cite{ding2021rfnet}. These fusion-based approaches adapt to modality-missing conditions by extracting modality-invariant and specific features but struggle to maintain performance in modality-complete conditions. As shown in Fig. \ref{motivation}(b) and (c), we apply modality dropout \cite{wei2023mmanet} to the baseline on RGB-T and RGB-D datasets respectively, resulting in a significant improvement in performance under modality-missing conditions. However, the performance under modality-complete conditions lags behind that of the baseline. 
To address this limitation, we explore a training strategy to enhance the performance of dual-modal SOD under both modality-complete and modality-missing conditions. Inspired by conditional controls \cite{zhang2023adding}, we treat the modality missing as the \textit{condition} and replicate the encoders trained when the modality is complete. Subsequently, we freeze the original encoder and update the parameters of the copied encoder and zero convolutions. This approach, namely Conditional Dropout (CD), preserves the model's capability when dealing with complete modalities and enables it to extract modality-invariant and specific features. 

We propose a new perspective on dual-modal SOD, aiming to investigate a robust framework resistant to noisy and incomplete inputs while maintaining accuracy when facing ideal inputs. In sum, the main contributions and why this work is non-trivial are as follows:
\begin{itemize}
    \item To the best of our knowledge, this is the first robust dual-modal SOD model against both noisy image and modality-missing.
    \item A language-driven quality assessment (LQA) module with a learnable prompt is proposed to reweight modality contributions. LQA enhances the robustness and performance of the model with noisy images.
    \item A Conditional Dropout (CD) learning method is proposed to promote the performance of the model in both modality complete and missing conditions.
    \item Though the proposed network is simply designed for better scalability, our model outperforms the state-of-the-art in both modality complete and missing conditions.
\end{itemize}

\section{Related Work}

\subsection{Dual-modal salient object detection}
Recent years have seen increased research in dual-modal salient object detection, particularly in RGB-T \cite{tu2022rgbt, tu2021multi, huo2021efficient, tu2022weakly, huo2022real, wang2021cgfnet, cong2022does, wang2023thermal, zhou2023lsnet} and RGB-D \cite{cheng2022depth, zhang2022c, cong2022cir, lee2022spsn, pang2023caver, liu2021visual, liu2024vst++, liu2020learning, liu2021learning} modalities.
% For example, \cite{huo2021efficient, huo2022real, tu2021multi, zhou2023lsnet} employ various fusion methods to enhance the interaction between the two modalities, thereby improving the model's dual-modal perception capabilities. However, they have not considered the potential issues of modality loss or deficiencies, which hinder their applicability in open-world scenarios. 
These works have made significant progress in the field of salient object detection.
% For instance, \cite{cong2022does, cheng2022depth} leveraged different estimation to enhance the performance of salient object detection, \cite{huo2021efficient} introduced context-guided fusion, \cite{huo2022real} proposed an early fusion single-stream network, \cite{tu2021multi} employed multi-interactive modules for fusion in the decoder phase, \cite{tu2022weakly} addressed the spatial misalignment issue between modalities, \cite{tu2022rgbt} selectively collected features from different branches and introduced an edge loss, \cite{wang2023thermal} compensated for low-illumination RGB images using thermal (T) images, \cite{wang2021cgfnet} introduced a cross-scale alternate guiding fusion approach, and \cite{zhou2023lsnet} focused on feature propagation between modalities. \cite{zhang2022c} developed a Criss-Cross Dynamic Filter Network, \cite{cong2022cir} emphasized cross-modality interaction and refinement, and \cite{lee2022spsn} proposed a Superpixel Prototype Sampling Network, all contributing to advancements in salient object detection.
For instance, \cite{cong2022does, cheng2022depth} enhanced performance through various estimations, \cite{huo2021efficient, huo2022real} focused on fusion methods like context-guided and early single-stream fusion, and \cite{tu2021multi, tu2022weakly} tackled issues like multi-interactive fusion and spatial misalignment. \cite{tu2022rgbt, wang2023thermal, wang2021cgfnet, zhou2023lsnet} introduced strategies like selective feature collection, compensation for low illumination, cross-scale fusion, and feature propagation. 
Additionally, \cite{zhang2022c, cong2022cir, lee2022spsn} contributed with methods like Criss-Cross Dynamic Filter Networks, cross-modality interaction, and Superpixel Prototype Sampling, respectively, to advance salient object detection.

These studies have collectively enhanced dual-modal salient object detection performance.
However, these methods frequently show limited adaptability to low-quality or absent modalities.
% Typically, models are trained under ideal situations, which restricts their generalization ability in scenarios where inputs are noisy or partially missing. 
Often, models trained under ideal conditions struggle with generalization in scenarios with noisy or partially missing inputs.
% We propose a model trained to handle both complete and incomplete modalities to overcome this limitation. 
To address this, we propose CoLA adept at handling both complete and incomplete modalities.
% This approach ensures robustness in scenarios where some modalities may be noisy or unavailable. 
CoLA guarantees robust performance in scenarios with noisy or unavailable modalities.

\subsection{Incomplete modality learning}
Incomplete modality learning aims to alleviate the modality missing problem due to sensor failures or environmental constraints \cite{ding2021rfnet}. 
Recent advantages can be categorized into reconstruction-based and fusion-based approaches. 
Reconstruction-based methods exploit the remaining modalities to recover or reconstruct the missing ones. 
For example, latent space strategies \cite{john2023multimodal} can be used to generate complete multimodal inputs, and Generative Adversarial Network \cite{cai2018deep, jue2019integrating, pan2020spatially} can be employed to restore the absent modalities. 
Reconstruction-based methods require training and deploying a distinct model for each subset of missing modalities, leading to high complexity in practical applications.
Recent works have explored fusion-based methods to learn joint representations directly from incomplete multimodal inputs across different applications. 
% \cite{ding2021rfnet} utilizes region-aware fusion networks for brain tumor segmentation in multimodal MRI with partially missing modalities.
% \cite{ma2022multimodal} employs multi-task optimization and a differentiable algorithm for incomplete modal data fusion.
% \cite{wei2023mmanet} uses knowledge distillation mechanisms for fusion in incomplete input cases.
% \cite{cui2022survival} combines fusion and reconstruction methods for brain cancer survival prediction with incomplete data.
% Such approaches avoid reconstruction errors but require designing effective fusion mechanisms. 
\cite{ding2021rfnet, ma2022multimodal, wei2023mmanet, cui2022survival} explore fusion methods for incomplete modalities from various perspectives such as regional perception, multi-tasking, and feature space.
% Existing methods for Dual-modal salient object detection have not considered solving the problem of missing modalities in dual-modal SOD, which limits the robustness of models in real-world applications. 

These methods tend to make the model adapt to the missing modality, which can hardly ensure unaffected performance on complete modalities. 
% Our proposed method enhances the model's feature extraction ability from each modality to address this issue.
We propose a Conditional Dropout method to improve performance in incomplete modalities without compromising complete modalities, alleviating the missing modality issue in dual-modal SOD tasks.

\subsection{Vision-language model and application}

Recent works research has extensively investigated a range of vision-language models, e.g., DALL-E \cite{ramesh2021zero}, ALIGN \cite{jia2021scaling} and CLIP \cite{radford2021learning}, which harmonizes text and image understanding by learning to associate images with their textual descriptions effectively.
These models have demonstrated remarkable capabilities across a spectrum of computer vision tasks such as crowd counting \cite{liang2023crowdclip}, pedestrian detection \cite{liu2023vlpd}, human and object interaction detection \cite{ning2023hoiclip}, scene text detection \cite{yu2023turning}, etc. 
Beyond recognition, CLIP is studied for monocular depth estimation in a zero-shot manner \cite{zhang2022can}. Due to the variety of downstream tasks, many fine-tuning improvements based on CLIP have been proposed \cite{zhou2022learning, zhou2022conditional, yu2023task}, which have greatly enhanced the versatility of CLIP \cite{wasim2023vita, yu2023turning, yu2023task}.
% Unsupervised approaches using CLIP \cite{radford2021learning} are proposed for crowd counting \cite{liang2023crowdclip} and pedestrian detection \cite{liu2023vlpd}. CLIP is also applied in supervised settings for HOI detection \cite{ning2023hoiclip}, image quality assessment \cite{zhang2023blind}, scene text detection \cite{yu2023turning}, etc. 
% Techniques like knowledge distillation and task residuals enable efficient tuning and few-shot learning \cite{yu2023task, gu2021open}.  

Zero-shot and fine-tune based CLIP applications present versatile approaches for diverse computer vision tasks.
We utilize CLIP to access the quality of inputs by prompt learning, leveraging its transformable capabilities to tackle the challenges posed by noisy inputs.
\begin{figure*}[t]
\centering
\includegraphics[width=0.9\textwidth,keepaspectratio]{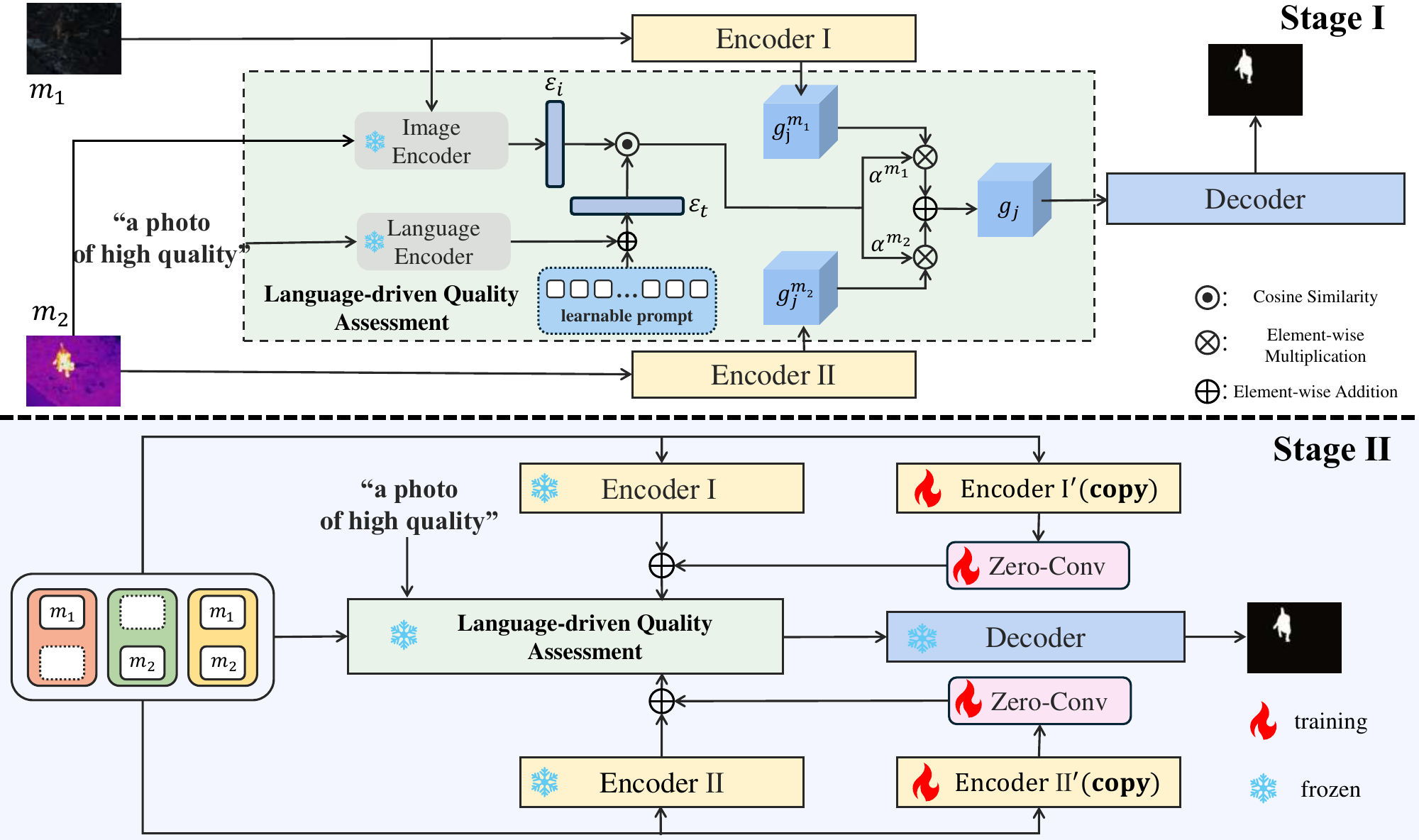}
\caption{The architecture of CoLA represents a two-stage neural network with Stage I training a language quality assessment (LQA) to calibrate feature fusion, and Stage II training with Conditional Dropout enhances the capabilities of both missing and complete modalities.}
\label{method}
\end{figure*}

\section{Methodology}
\subsection{Overview of the framework}

The process of training a conventional multi-modal SOD can be summarized as $f(x) \stackrel{t_{1}}{\leftarrow}\left\{f;x\right\}$ where $t_1$ denotes the training process, $x$ refers to the multi-modal inputs, and $f(x)$ is the trained model with data $x$.

Unlike previous works, our CoLA consists of four components: 1) a dual-branch encoder for feature extraction; 2) an identically structured encoder with a Conditional Dropout ($CD$) for robust modality feature supplementation (in training stage \uppercase\expandafter{\romannumeral 2}); 3) a Language-driven Quality Assessment module (LQA) for learning modality contribution ratios (in training stage \uppercase\expandafter{\romannumeral 1}); 4) a decoder for feature aggregation to produce the output. Importantly, our encoder and decoder designs are kept simple without complex interactions. This allows us to demonstrate the efficacy of our proposed modules clearly.
% Let us denote the input as $x$, the first stage training as $f$, the second stage training as $f^{\prime}$, and the output as $y$. Based on these definitions, 
% For our proposed method, the formulation of the first stage can be expressed as $f(x) \stackrel{t_{1}}{\leftarrow}\left\{f^{\prime}; x\right\}$ and the second stage can be expressed as $f(x) \stackrel{t_{2}}{\leftarrow}\left\{f^{\prime}; L Q A; CD; \rho(x)\right\}$. 
Based on the components, the training process can be expressed as:
\begin{equation}
    f^{\prime}(x) \xleftarrow{t_{1}} \{f^{\prime};LQA(x)\},
\end{equation}
\begin{equation}
    f(x) \xleftarrow{t_{2}} \{f^{\prime}(\rho(x)); CD(\rho(x))\},
\end{equation}
where $x$ is the multi-modal inputs, $f^{\prime}(x)$ represents the trained model of the first stage. The first and second stage training processes are denoted by $t_1$ and $t_2$ respectively, and $\rho(x)$ is the input with missing probability $\rho$.
% the probability of missing values in the two input modalities.
% In this formulation, the output of the first stage $f(x)$ is further optimized in the second stage using $f'$ together with the proposed $CD$ method, conditioned on the predicted quality scores $\rho(x)$. 
% The detailed workflow is elaborated in Section \ref{Learning objective}.
\begin{figure}[t]
\centering
\includegraphics[width=0.9\textwidth,keepaspectratio]{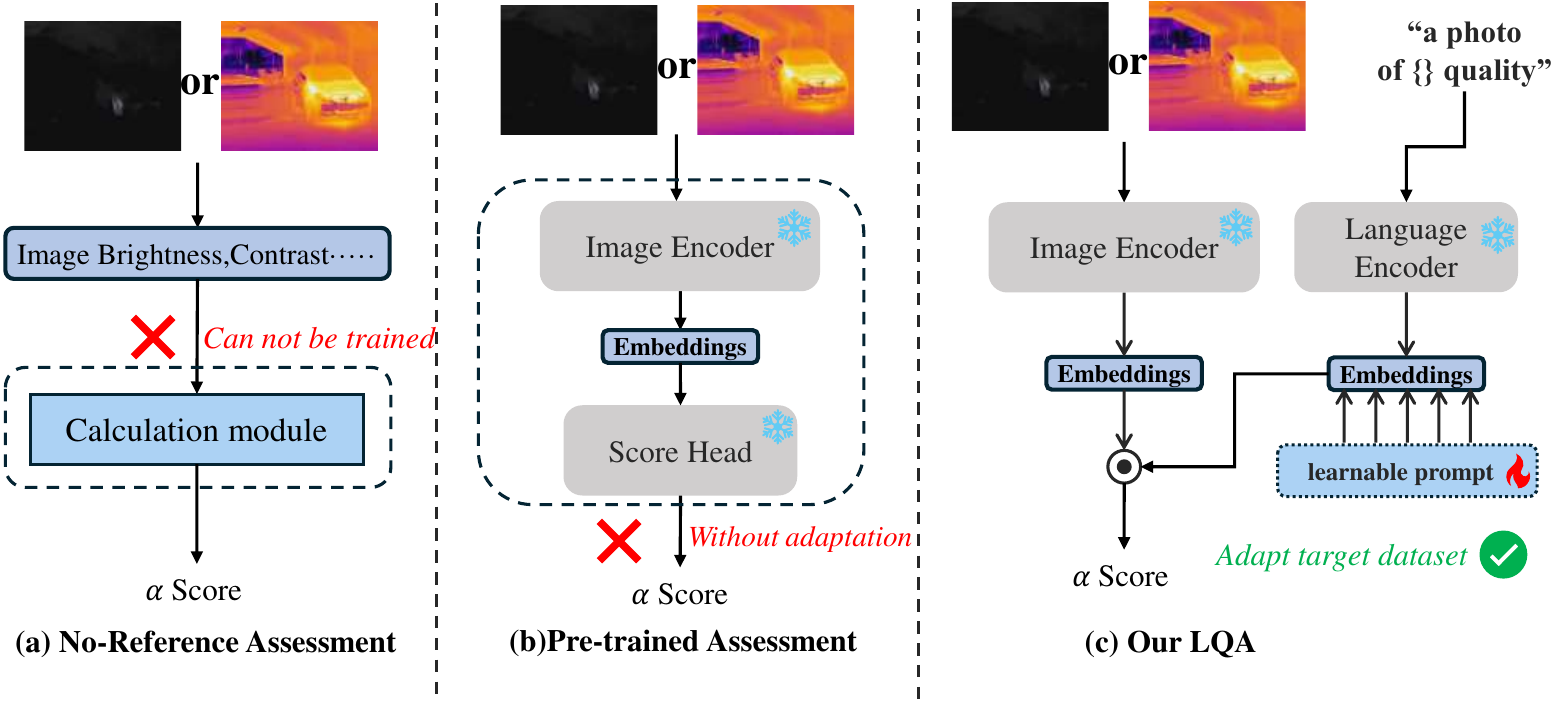}
\caption{Architectural comparison of (a) No-Reference Method, (b) Pre-trained Assessment and (c) Our LQA.}
\label{fig_compare_LQA}
\end{figure}
\subsection{Language-driven quality assessment for low quality image}
% 图像加名字，特征改名g
% Corresponding to the robustness of modality quality (contribution).
%思路:先说我们准备用α来衡量模态的贡献,从而使模型能够充分学习到模态共有的特征.接下来说α是怎么来的,然后介绍α是如何工作的(怎么控制融合),最后概括方法
% The original CLIP is not designed for modality quality assessment. It is worthwhile studying how to adjust it to our framework adaptively. 
% We adopt the text prompt "a photo of {} quality" and the image in the LQA module as the input. 
Fig. \ref{fig_compare_LQA} compares other image quality assessment methods with LQA. Existing assessment methods can be mainly categorized into two types. The first type (Fig. \ref{fig_compare_LQA}(a)) relies on the inherent characteristics of the image, such as brightness, contrast, and other attributes, to calculate the image quality,  such as BRISQUE \cite{mittal2012no}.
The second type uses neural network models pre-trained on other dataset (Fig. \ref{fig_compare_LQA}(b)), for example, GIE\cite{cong2022does}. 
The first type is constrained by its incapacity for training, while the second type can not adapt to the target dataset. Inspired by CLIP-IQA \cite{wang2023exploring}, we found vision-language model have demonstrated superior performance in assessing the quality of an image, so we design LQA which based on fine-tuning vision-language model (Fig. \ref{fig_compare_LQA}(c)) for modality quality assessment to reweight the contribution of each modality to extract robust features from dual-modal inputs. The process of LQA is formulated as follows:
\begin{equation}
    g = LQA(\mathcal{M}, \mathcal{F}(\mathcal{M}; \theta), \mathcal{T}),
\end{equation}
where $\mathcal{M} = \{ m_1, m_2 \}\in \mathbb{R}^{3\times H\times W}$ are a pair of dual-modal images, e.g., RGB and Thermal in Fig. \ref{method}. $\mathcal{T}$ are texts; $\mathcal{F}$ is the encoder \uppercase\expandafter{\romannumeral 1} and \uppercase\expandafter{\romannumeral 2} with parameters $\theta$; $g$ are the output of LQA and fed into the decoder. We first pass $m_1$ and $m_2$ through the image encoder of CLIP to obtain the image embedding $\varepsilon_i \in \mathbb{R}^{1\times D} $, where $D$ is the image embedding dimension ($D$ is set to 512). In parallel, the text encoder in CLIP receives inputs $\mathcal{T}$$=\{A\ photo \ of\  high\  quality.\}$ and outputs text embeddings $\varepsilon_t \in \mathbb{R}^{1\times D}$. To adapt the target dataset, we add a learnable prompt $\omega$ to the text embedding $\varepsilon_t$:
\begin{equation}
    \varepsilon_i = \varepsilon_i + \omega.
\end{equation}
In this process, we explore adapting pre-trained vision-language models for quality assessment by adding a small trainable parameter to the text encoder. The parameter $\omega$ is trained to enhance the model's robustness against noisy images. To this end, we define the quality score estimated by CLIP on the two modality images as $\alpha^{m_1},\alpha^{m_2}$:
\begin{gather}\label{alpha}
    \alpha^{m_1} = sim(\varepsilon_i^{m_1}, \varepsilon_t) = \frac{\varepsilon_i^{m_1}\cdot \varepsilon_t}{||\varepsilon_i|| \ ||\varepsilon_t||},\   
    \\
    \alpha^{m_2} = sim(\varepsilon_i^{m_2}, \varepsilon_t) = \frac{\varepsilon_i^{m_2}\cdot \varepsilon_t}{||\varepsilon_i|| \ ||\varepsilon_t||},\
\end{gather}
where $\varepsilon_i^{m_1}$ and $\varepsilon_i^{m_2}$ are the image embeddings from the CLIP image encoder,
$sim$ indicates cosine similarity between $\varepsilon_i$ and $\varepsilon_t$. It should be noted that LQA is trained only in stage \uppercase\expandafter{\romannumeral 1}, which is expected to improve the ability to recognize noisy images. Its parameters remain fixed and are not updated in the training stage \uppercase\expandafter{\romannumeral 2}. Subsequently, we fuse the individual layers of RGB and T features using the $\alpha$:
\begin{equation}\label{alpha_fusion}
    g_j = g_j^{m_1}* \frac{\alpha^{m_1}}{\alpha^{m_1}+\alpha^{m_2}} + g_j^{m_2}*\frac{\alpha^{m_2}}{\alpha^{m_1}+\alpha^{m_2}},
\end{equation}
where $g_j^{m_1}$ and $g_j^{m_2}$ represent the $j$-th layer features of the first and second modality respectively; $g = \{g_1, ..., g_n\}$ is the fused feature passed into the decoder; $\alpha^{m_1}$ and $\alpha^{m_2}$ are quality score for ${m_1}$ and ${m_2}$. For the decoder, we perform standard CBAM operations on $g_j$, then conv and upsample before adding to $g_{j+1}$. Based on our model, $n$ is set to 5.

\subsection{Conditional Dropout for incomplete multi-modal learning}
\begin{figure}[t]
\centering
\includegraphics[width=0.45\textwidth,keepaspectratio]{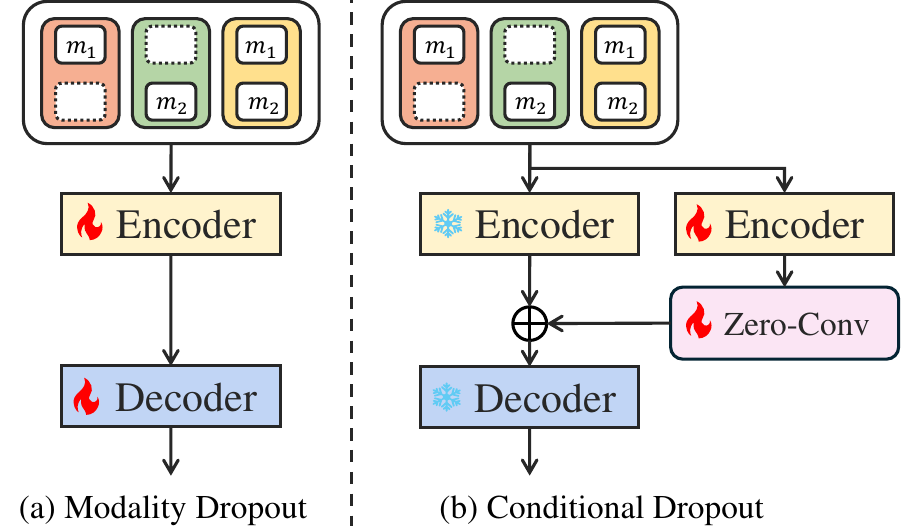}
\caption{Architectural comparison of (a) modality dropout \cite{wei2023mmanet} and (b) Conditional Dropout in dual-modal object detection.}
\label{fig_compare}
\end{figure}
% 我们的方法是怎么做的:dropout+conditional control
Previous methods for handling missing modalities mostly employ direct modality dropout \cite{wei2023mmanet} (Fig. \ref{fig_compare}(a)) during training, which can improve model performance with missing modalities but significantly degrade performance with complete modalities. We proposed Conditional Dropout (Fig. \ref{fig_compare}(b)) to preserve the capability in modality-complete scenarios by freezing the pre-trained model and finetuning the trainable copy and zero convolution.
 
% Specifically, in the first training stage, we utilize complete modalities $m_1$ and $m_2$ to establish perceptual abilities in the two modalities: 
% \begin{equation}
%     y = \mathcal{F}(\mathcal{M}; \theta),
% \end{equation}
In stage \uppercase\expandafter{\romannumeral 2}, we will choose one of the input pairs $\rho(\mathcal{M})$ from the following conditions:
% each input modality $m_1$ and $m_2$  will be assigned as a condition by probability before being fed into the encoder: 
% \begin{equation}
%     \rho(\mathcal{M}) = \left\{
%     \begin{alignedat}{2}
%         Cond_1(\mathcal{M})\triangleq &\{m_1, m_2\}\\
%         Cond_2(\mathcal{M})\triangleq &\{m_1, \  \phi \ \ \}\\
%         Cond_3(\mathcal{M})\triangleq &\{\ \phi\ \ , m_2 \}
%     \end{alignedat}
%     \right.
% \end{equation}
% \begin{equation}
%     \rho(\mathcal{M}) = \left\{
%     \begin{alignedat}{2}
%         &\{m_1, m_2\},\ \ \ \ \ \ \  \rho_1 \coloneqq \\
%         &\{m_1, \  \phi \ \ \},\ \ \ \ \ \ \ \rho_2\\
%         &\{\ \phi\ \ , m_2 \},\ \ \ \ \ \ \ \rho_3
%     \end{alignedat}
%     \right.
% \end{equation}
\begin{equation}
    \rho(\mathcal{M}) = \left\{
    \begin{alignedat}{2}
        Cond_1 := \  &\{m_1, m_2\}\\
        Cond_2 := \  &\{m_1, \  \phi \ \ \}\\
        Cond_3 := \  &\{\ \phi\ \ , m_2 \}
    \end{alignedat}
    \right.,
\end{equation}
where $\phi$ denotes the zero input, $Cond_1$, $Cond_2$, and $Cond_3$ denote the selection probabilities for three distinct input pairs. 
% $Cond_1$ denotes modality-complete input, while $Cond_2$ and $Cond_3$ denote modality-missing inputs.
The original parameters $\theta$ of the encoder from the stage  \uppercase\expandafter{\romannumeral 1} are frozen and we duplicate the encoder to create a trainable copy Encoder ${\rm \uppercase\expandafter{\romannumeral 1}^\prime}$, Encoder ${\rm \uppercase\expandafter{\romannumeral 2}^\prime}$ with parameters $\theta_f$. Then we connect them using zero convolutions $\mathcal{Z}$ with both weight and bias initialized to zero:
\begin{equation}\label{cond_drop}
    g = \mathcal{F}(\rho(\mathcal{M}); \theta) + \mathcal{Z}(\mathcal{F}(\rho(\mathcal{M}); \theta_f); \theta_z),
\end{equation}
where $\theta_z$ is the parameter of the zero convolution.
% This conditional dropout approach maintains the perception of full modalities while enhancing single-modality robustness.
% 我们方法优势在哪里,效果好
%The proposed conditional dropout method retains the perception of full modalities while enhancing robustness to incomplete cases. It maintains representation quality on complete data by preventing interference with the original encoder. Experiments validate that our technique performs better on full modalities than conventional dropout, enhancing robustness to missing modalities. 
According to Eq. \ref{cond_drop}, the first term $\mathcal{F}(\rho(\mathcal{M}); \theta)$ represents the frozen model $f$ trained in stage I with parameter $\theta$, which gains the ability to recognize common features from dual modalities. The second term $\mathcal{Z}(\mathcal{F}(\rho(\mathcal{M}); \theta_f); \theta_z)$ denotes the additional parameters $\theta_f$ and $\theta_z$ introduced by the Conditional Dropout in the second stage. By using incomplete modalities as input during training, we force the model to extract more fine-grained representations from each modality. In summary, the single modality inputs compel the model to learn richer representations recorded in the second term. Meanwhile, the frozen first term preserves the previously learned knowledge. Finally, the $\mathcal{Z}(\cdot;\cdot)$ with zero initialization ensures minimal influence at the beginning of the training stage \uppercase\expandafter{\romannumeral 2}, and the newly learned enriched features are progressively integrated into the model during training.
The Conditional Dropout framework paves a way to obtain missing data robustness while maintaining accuracy on complete inputs. This framework could strengthen both single-modality and full-modality potentials concurrently.
% The conditional control framework presents a promising direction for obtaining missing data robustness without sacrificing accuracy on complete inputs.
\definecolor{myPink}{RGB}{255 ,105 ,180}
\subsection{Learning objective}\label{Learning objective}
% loss function.\\
% training steps.
Given a pair of images $m_1, m_2$ and an input prompt $\mathcal{T}$, in the training stage \uppercase\expandafter{\romannumeral 1}, we input $m_1$ and $\mathcal{T}$ into the proposed LQA module, learning to predict scores $\alpha$ for evaluating image quality as shown in Eq. \ref{alpha}. The $m_1$ and $m_2$ are also fed into the encoder, and the output $g_i^{m_i}$ are fused by $\alpha$ in Eq. \ref{alpha_fusion}. After passing through the decoder, the loss is calculated between the output and GT. During training, We use BCE loss and IOU loss as our loss functions:
\begin{equation}\label{loss}
    \mathcal{L}_{total} = \mathcal{L}_{bce}(pred, GT) + \mathcal{L}_{iou}(pred, GT).
\end{equation}

% In the training stage \uppercase\expandafter{\romannumeral 2}, we freeze the parameters of the previously trained encoder and initialize the trainable copy and simultaneously feed the images into the encoder from stage \uppercase\expandafter{\romannumeral 1} and the new encoders built in the stage \uppercase\expandafter{\romannumeral 2} in Eq. \ref{cond_drop}. 
During training stage \uppercase\expandafter{\romannumeral 2}, parameters of the initial encoder are frozen, a trainable copy is initialized, and images are fed into both the stage \uppercase\expandafter{\romannumeral 1} encoder and the new stage \uppercase\expandafter{\romannumeral 2} encoders as per Eq. \ref{cond_drop}.
% The output of $Encoder \uppercase\expandafter{\romannumeral 1}^{\prime}$ and $Encoder \uppercase\expandafter{\romannumeral 2}^{\prime}$ is fed into the zero-initialized zero convolution $\mathcal{Z}(\cdot;\cdot)$ so that the model will not be interfered by the noise generated from it at the beginning of optimization. Then, we add the output of the $\mathcal{Z}(\cdot;\cdot)$ and the original encoder's output before feeding into the decoder. 
After feeding into the decoder, the loss is calculated between the output and GT to optimize the parameters of the stage \uppercase\expandafter{\romannumeral 2} trainable copy and zero convolution.
CoLA focuses on simplicity and extensibility rather than complex model design. The encoder, decoder, and training loss are kept straightforward without meticulous engineering. 
Combining the above methods, our approach establishes a new evaluation in the dual-modal SOD field, capable of maintaining the model's accuracy under conditions of noisy and missing inputs.
\section{Experiments}
\definecolor{myRed}{RGB}{219, 68, 55}
\definecolor{myGreen}{RGB}{15, 157, 88}
\definecolor{myBlue}{RGB}{66, 133, 244}
\definecolor{myPink}{RGB}{255 ,105 ,180}
\label{sec:exp}
\subsection{Experiment setup}

\textbf{Dataset.} For RGB-T, we employ three widely recognized public datasets for our experiment: VT821 \cite{wang2018rgb}, VT1000 \cite{tu2020rgbt}, and VT5000 \cite{tu2022rgbt}. In terms of training, we utilized 2500 image pairs from the VT5000 dataset for training purposes. The model’s performance under modality-complete and modality-missing was evaluated using the remaining 2500 image pairs from the VT5000 dataset, along with the VT821 and VT1000 datasets. \\For RGB-D, we utilize four well-known public datasets for our experiment: SIP \cite{fan2020rethinking}, NJUK \cite{ju2014depth}, DES \cite{cheng2014depth}, and NLPR \cite{peng2014rgbd}. For training purposes, we used a combination of 1485 image pairs from the NJUK dataset and 700 image pairs from the NLPR dataset.\\\\
% For data augmentation, we employ various strategies such as random flipping, rotation, boundary clipping, adding probabilistic noise, and multi-scale input.
\textbf{Implementation details. }We use a single NVIDIA GeForce RTX 3090 GPU. The backbone of all methods utilizes a ResNet-50 \cite{he2016deep} model pre-trained on the ImageNet \cite{deng2009imagenet} dataset. We use the Adam optimizer with an initial learning rate of 1e-4. In stage \uppercase\expandafter{\romannumeral1}, we train the model for a total of 100 epochs with a batch size of 8 and divide the learning rate by 10
every 45 epochs. In stage \uppercase\expandafter{\romannumeral2}, we conduct a training period covering 60 epochs, starting with an initial learning rate set at 1e-4. This learning rate is then decreased by a factor of 10 after every 35 epochs.\\
\begin{table*}[t]
\caption{Quantitative experiments of different RGB-T models in modality-complete and modality-missing conditions. $\uparrow$ denotes that a larger value is better. The best results are in red, the second-best results are in blue, and the third-best results are in green.}
% \label{pe-c-surf}
\centering
\label{table_comparison with others}
\fontsize{6.5}{9}\selectfont
\begin{tabular}{c|p{0.8cm}<{\centering}p{0.6cm}<{\centering}|c|*{6}{p{0.9cm}<{\centering}}p{1.1cm}<{\centering}|c}
\toprule
    \multirow{2}{*}{Datasets} & \multicolumn{2}{c|}{Conditions} & \multirow{2}{*}{Metric}  & CSRNet   & ADF & TNet &DCNet & LSNet &MIDD     &TAGFNet      & Ours\\
\cmidrule{2-3}         & \multicolumn{1}{c|}{RGB} & T     &          & \cite{huo2021efficient}  & \cite{tu2022rgbt} &\cite{cong2022does} &\cite{tu2022weakly} &\cite{zhou2023lsnet} &\cite{tu2021multi}    &\cite{wang2023thermal} &\\
    \midrule
     \multirow{10}{*}{VT821} & \multirow{2}[1]{*}{\solidcircle} & \multirow{2}[1]{*}{\solidcircle} & 
    \textit{$E_m$$\uparrow$}  & 0.909 & 0.841 & \textcolor{myBlue}{0.919} & 0.911 & 0.910 & 0.901  &\textcolor{myGreen}{0.912}&  \textcolor{myRed}{\textbf{0.922}} \\
  &       &       & 
    \textit{$F_\beta$$\uparrow$}  & \textcolor{myGreen}{0.837} & 0.718 & 0.823 & \textcolor{myBlue}{0.841} & 0.829 & 0.819 &0.825& \textcolor{myRed}{\textbf{0.849}}    \\
  & \multirow{2}[0]{*}{\emptycircle} & \multirow{2}[0]{*}{\solidcircle} & 
    \textit{$E_m$$\uparrow$}  & 0.720 & 0.640 & \textcolor{myGreen}{0.834} & 0.750 & 0.787 & 0.799 &\textcolor{myBlue}{0.855}&  \textcolor{myRed}{\textbf{0.864}} \\
  &       &       & 
    \textit{$F_\beta$$\uparrow$}  & 0.540 & 0.465 & \textcolor{myGreen}{0.713} & 0.644 & 0.687 & 0.680 &\textcolor{myBlue}{0.727} &\textcolor{myRed}{\textbf{0.752}}  \\
  & \multirow{2}[1]{*}{\solidcircle} & \multirow{2}[1]{*}{\emptycircle} 
   & 
    \textit{$E_m$$\uparrow$}  & 0.726 & 0.765 & 0.813 & \textcolor{myBlue}{0.872} & 0.841 & 0.840  &\textcolor{myGreen}{0.861}&\textcolor{myRed}{\textbf{0.900}}     \\
  &       &       & 
    \textit{$F_\beta$$\uparrow$}  & 0.551 & 0.655 & 0.740 & 0.780 & 0.749 & 0.737  &0.771&  \textcolor{myRed}{\textbf{0.817}}
  \\
\cmidrule{2-12}          & \multicolumn{2}{c|}{\multirow{2}{*}{Average Drop}} & 
    \textit{\(E_m\)\(\uparrow\)}  & -0.186 & -0.139 & -0.096 & -0.100 & -0.096 & \textcolor{myGreen}{-0.082}  &\textcolor{myBlue}{-0.049} & \textcolor{myRed}{\textbf{-0.040}}     \\
  & \multicolumn{2}{c|}{} & 
    \textit{\(F_\beta\)\(\uparrow\)} & -0.291 & -0.158 & \textcolor{myGreen}{-0.097} & -0.129 & -0.111 & -0.110  &\textcolor{myBlue}{-0.076}&   \textcolor{myRed}{\textbf{-0.065}} \\       
    
    & \multicolumn{2}{c|}{\multirow{2}{*}{Average}} & 
    \textit{$E_m$$\uparrow$}  & 0.785 & 0.749 & \textcolor{myGreen}{0.855} & 0.844 & 0.846 & 0.847  &\textcolor{myBlue}{0.876}&  \textcolor{myRed}{\textbf{0.895}}     \\
  & \multicolumn{2}{c|}{} & 
    \textit{$F_\beta$$\uparrow$} & 0.643 & 0.613 & \textcolor{myGreen}{0.759} & 0.755 & 0.755 & 0.745  &\textcolor{myGreen}{0.774}&  \textcolor{myRed}{\textbf{0.806}} \\
    \midrule

    % VT1000数据集数据
    \multirow{10}{*}{VT1000} & \multirow{2}[1]{*}{\solidcircle} & \multirow{2}[1]{*}{\solidcircle} & 
      \textit{$E_m$$\uparrow$}  & \textcolor{myBlue}{0.952} & 0.928 & 0.937 & 0.949 & \textcolor{myGreen}{0.951} & 0.946&\textcolor{myBlue}{0.952}&    \textcolor{myRed}{\textbf{0.955}}    \\
      &       &       & 
      \textit{$F_\beta$$\uparrow$}  & \textcolor{myGreen}{0.900} & 0.873 & 0.885 & \textcolor{myBlue}{0.902} & 0.882 & 0.884&0.890 & \textcolor{myRed}{\textbf{0.904}}    \\
      & \multirow{2}[0]{*}{\emptycircle} & \multirow{2}[0]{*}{\solidcircle} & 
      \textit{$E_m$$\uparrow$}  & 0.789 & 0.761 & \textcolor{myGreen}{0.896} & 0.788 & 0.858 & 0.886&\textcolor{myBlue}{0.919}  &   \textcolor{myRed}{\textbf{0.930}}   \\
      &       &       & 
      \textit{$F_\beta$$\uparrow$}  & 0.634 & 0.656 & \textcolor{myGreen}{0.810} & 0.723 & 0.783 & 0.807&\textcolor{myBlue}{0.834}&  \textcolor{myRed}{\textbf{0.855}}   \\
      & \multirow{2}[1]{*}{\solidcircle} & \multirow{2}[1]{*}{\emptycircle} 
      & 
    \textit{$E_m$$\uparrow$}  & 0.781 & 0.872 & 0.891 & \textcolor{myRed}{\textbf{0.943}} & 0.913 & \textcolor{myGreen}{0.927}&\textcolor{myBlue}{0.936}   &\textcolor{myRed}{\textbf{0.943}}    \\
          &       &       & 
      \textit{$F_\beta$$\uparrow$}  & 0.652 & 0.801 & 0.839 & \textcolor{myBlue}{0.884} & 0.853 & 0.860&\textcolor{myGreen}{0.871}    &\textcolor{myRed}{\textbf{0.887}}    
      \\
\cmidrule{2-12}          & \multicolumn{2}{c|}{\multirow{2}{*}{Average Drop}} & 
    \textit{\(E_m\)\(\uparrow\)}  & -0.167 & -0.112 & -0.044 & -0.083 & -0.065 & \textcolor{myGreen}{-0.040}  &\textcolor{myBlue}{-0.025} & \textcolor{myRed}{\textbf{-0.018}}     \\
  & \multicolumn{2}{c|}{} & 
    \textit{\(F_\beta\)\(\uparrow\)} & -0.257 & -0.145 & -0.061 & -0.099 & -0.064 & \textcolor{myGreen}{-0.050}  &\textcolor{myBlue}{-0.038}&  \textcolor{myRed}{\textbf{-0.027}} \\
      & \multicolumn{2}{c|}{\multirow{2}{*}{Average}} & 
      \textit{$E_m$$\uparrow$}  & 0.841 & 0.854 & 0.908 & 0.893 & 0.907 & \textcolor{myGreen}{0.920}& \textcolor{myBlue}{0.936}  &  \textcolor{myRed}{\textbf{0.943}}   \\
      & \multicolumn{2}{c|}{} & 
      \textit{$F_\beta$$\uparrow$}  & 0.729 & 0.777 & 0.845 & 0.836 & 0.839 & \textcolor{myGreen}{0.851}& \textcolor{myBlue}{0.865}&  \textcolor{myRed}{\textbf{0.882}}   \\
\midrule
% VT821数据集数据.
\multirow{10}{*}{VT5000} & \multirow{2}[1]{*}{\solidcircle} & \multirow{2}[1]{*}{\solidcircle} & 
    \textit{$E_m$$\uparrow$}  & 0.901 & 0.887 & \textcolor{myRed}{\textbf{0.927}} & \textcolor{myBlue}{{0.922}} & \textcolor{myGreen}{0.914} & 0.906&0.913 &\textcolor{myRed}{\textbf{0.927}}    \\
          &       &       & 
    \textit{$F_\beta$$\uparrow$}  & 0.818 & 0.802 & \textcolor{myBlue}{0.827} & \textcolor{myGreen}{0.822} & 0.820 & 0.807&0.819    & \textcolor{myRed}{\textbf{0.843}} \\
    & \multirow{2}[0]{*}{\emptycircle} & \multirow{2}[0]{*}{\solidcircle} & 
    \textit{$E_m$$\uparrow$}  & 0.744 & 0.751 & \textcolor{myGreen}{0.854} & 0.693 & 0.801 & 0.819&\textcolor{myBlue}{0.869}  &\textcolor{myRed}{\textbf{0.887}} \\
          &       &       & 
    \textit{$F_\beta$$\uparrow$}  & 0.523 & 0.603 & \textcolor{myBlue}{0.743} & 0.572 & 0.689 & 0.699&\textcolor{myGreen}{0.742}   &\textcolor{myRed}{\textbf{0.774}}\\
    
    & \multirow{2}[1]{*}{\solidcircle} & \multirow{2}[1]{*}{\emptycircle} 
    & 
    \textit{$E_m$$\uparrow$}   & 0.733 & 0.820  & 0.786 & \textcolor{myGreen}{0.895} & 0.847 & 0.874&\textcolor{myGreen}{0.895}  &\textcolor{myRed}{\textbf{0.913}}\\ &       &       & 
    \textit{$F_\beta$$\uparrow$}   & 0.550  & 0.707 & 0.717 & \textcolor{myBlue}{0.792} & 0.752 & 0.764&\textcolor{myGreen}{0.791}  &\textcolor{myRed}{\textbf{0.822}}\\
\cmidrule{2-12}          & \multicolumn{2}{c|}{\multirow{2}{*}{Average Drop}} & 
    \textit{\(E_m\)\(\uparrow\)}  & -0.163 & -0.102 & -0.107 & -0.128 & -0.090 & \textcolor{myGreen}{-0.060}  &\textcolor{myBlue}{-0.031} & \textcolor{myRed}{\textbf{-0.027}}     \\
  & \multicolumn{2}{c|}{} & 
    \textit{\(F_\beta\)\(\uparrow\)} & -0.281 & -0.147 & -0.097 & -0.140 & -0.100 & \textcolor{myGreen}{-0.076}  &\textcolor{myBlue}{-0.052}&  \textcolor{myRed}{\textbf{-0.045}} \\
          & \multicolumn{2}{c|}{\multirow{2}{*}{Average}} &
    \textit{$E_m$$\uparrow$}  & 0.793 & 0.819 & 0.856 & 0.837 & 0.854 & \textcolor{myGreen}{0.866}&\textcolor{myBlue}{0.892} &\textcolor{myRed}{\textbf{0.909}}     \\
          & \multicolumn{2}{c|}{} & 
          \textit{$F_\beta$$\uparrow$}  & 0.630 & 0.704 & \textcolor{myGreen}{0.762} & 0.729 & 0.754 & 0.757&  \textcolor{myBlue}{0.784}& \textcolor{myRed}{ \textbf{0.813}}\\
    \bottomrule
\end{tabular}

\end{table*}
\\
\textbf{Evaluation metrics.} We employ four widely used evaluation metrics to validate the performance of our model and other existing SOTA methods: S-measure($S_\alpha$) \cite{fan2017structure}, Mean F-measure ($F_\beta$) \cite{achanta2009frequency}, Mean E-measure($E_m$) \cite{fan2018enhanced} and Mean Square Error(MAE) \cite{borji2015salient}. \textbf{Average Drop} and \textbf{Average}, to measure the robustness and overall capability of a dual-modal SOD model respectively. \textbf{Average Drop} refers to the average performance drop when modalities are missing compared to when modalities are complete. \textbf{Average} refers to the average performance across the three conditions. \\
 
\subsection{Compare with state-of-the-art}
To validate the effectiveness of our CoLA in RGB-T SOD under modality-complete and modality-missing, we compared it with nine state-of-the-art methods from the past three years. The compared methods include ADF \cite{tu2022rgbt}, MIDD \cite{tu2021multi}, CSRNet \cite{huo2021efficient}, DCNet \cite{tu2022weakly}, TNet \cite{cong2022does}, TAGFNet \cite{wang2023thermal} and LSNet \cite{zhou2023lsnet}.
We also compared our CoLA with six state-of-the-art methods in RGB-D from the past three years. The compared methods include D3Net \cite{fan2020rethinking}, DIGR \cite{cheng2022depth}, C$^{2}$DFNet \cite{zhang2022c}, CIRNet \cite{cong2022cir}, SPSN \cite{lee2022spsn}, HiDAnet\cite{wu2023hidanet}.

Our quantitative comparison results are shown in Table \ref{table_comparison with others}. The compared models can be classified into two categories: models like CSRNet \cite{huo2021efficient} and ADF \cite{tu2022rgbt} exhibit significant performance degradation when either RGB or Thermal modality is missing, indicating poor robustness in handling modality-missing conditions. Some methods like DCNet\cite{tu2022weakly} perform well when one modality is missing but suffer greatly when the other is absent, indicating an excessive reliance on a particular modality. 

Our method excels by achieving the best results in both Average Drop and Average across all datasets, demonstrating unparalleled robustness and overall performance. CoLA also achieved the best performance when dealing with modality-complete as well as when dealing with modality-missing.
\begin{table*}[t]
\caption{Ablation experiments for each component on the VT5000 dataset. The Baseline is the network with a value of $\alpha$ set to 0.5.}
\centering
\fontsize{5.5}{8}\selectfont
\begin{tabular}{p{0.2cm}<{\centering}p{2cm}|p{0.4cm}<{\centering}p{0.4cm}<{\centering}p{0.4cm}<{\centering}p{0.6cm}<{\centering} |*{3}{p{0.4cm}}<{\centering}p{0.6cm}<{\centering}|*{3}{p{0.4cm}}<{\centering}p{0.6cm}<{\centering}|*{3}{p{0.4cm}}<{\centering}p{0.6cm}<{\centering}}
\toprule
&\multirow{2}{*}{}& \multicolumn{4}{c|}{Modality Complete} & \multicolumn{4}{c|}{Missing RGB} & \multicolumn{4}{c|}{Missing Thermal} & \multicolumn{4}{c}{Average} \\
 &&$S_\alpha$$\uparrow$ & $E_m$$\uparrow$ & $F_\beta$$\uparrow$ & MAE$\downarrow$ & $S_\alpha$$\uparrow$ & $E_m$$\uparrow$ & $F_\beta$$\uparrow$ & MAE$\downarrow$ & $S_\alpha$$\uparrow$ & $E_m$$\uparrow$ & $F_\beta$$\uparrow$ & MAE$\downarrow$ & $S_\alpha$$\uparrow$ & $E_m$$\uparrow$ & $F_\beta$$\uparrow$ & MAE$\downarrow$ \\
\midrule
(a)&Baseline &.859 &.892 &.801 &.052 &.820 &.866 &.727 &.064 &.845& .883&.793 &.058 &.841 &.880 &.774 &.058 \\
(b)&Baseline+LQA &.887 &.923 &.834 &.039 &.828 &.876 &.754 &.059 &.849& .884&.789 &.061 &.855 &.894 &.792 &.053 \\
(c)&Baseline+CD &.880 &.908 &.818 &.043 &.833 &.870 &.750 &.061 &.868& .902 &.811 &.044 &.860 &.892 &.793 &.049 \\
(d)&Baseline+LQA+CD&\textbf{.892} &\textbf{.927} &\textbf{.843} &\textbf{.037} &\textbf{.840} &\textbf{.887} &\textbf{.774} &\textbf{.052} &\textbf{.874} &\textbf{.913} &\textbf{.822} &\textbf{.042} &\textbf{.869} &\textbf{.909} &\textbf{.813} &\textbf{.044}  \\
\bottomrule
\end{tabular}
\label{table_Ablation for all}
\end{table*}
\begin{figure*}[t]
\centering
\includegraphics[width=0.95\linewidth,keepaspectratio]{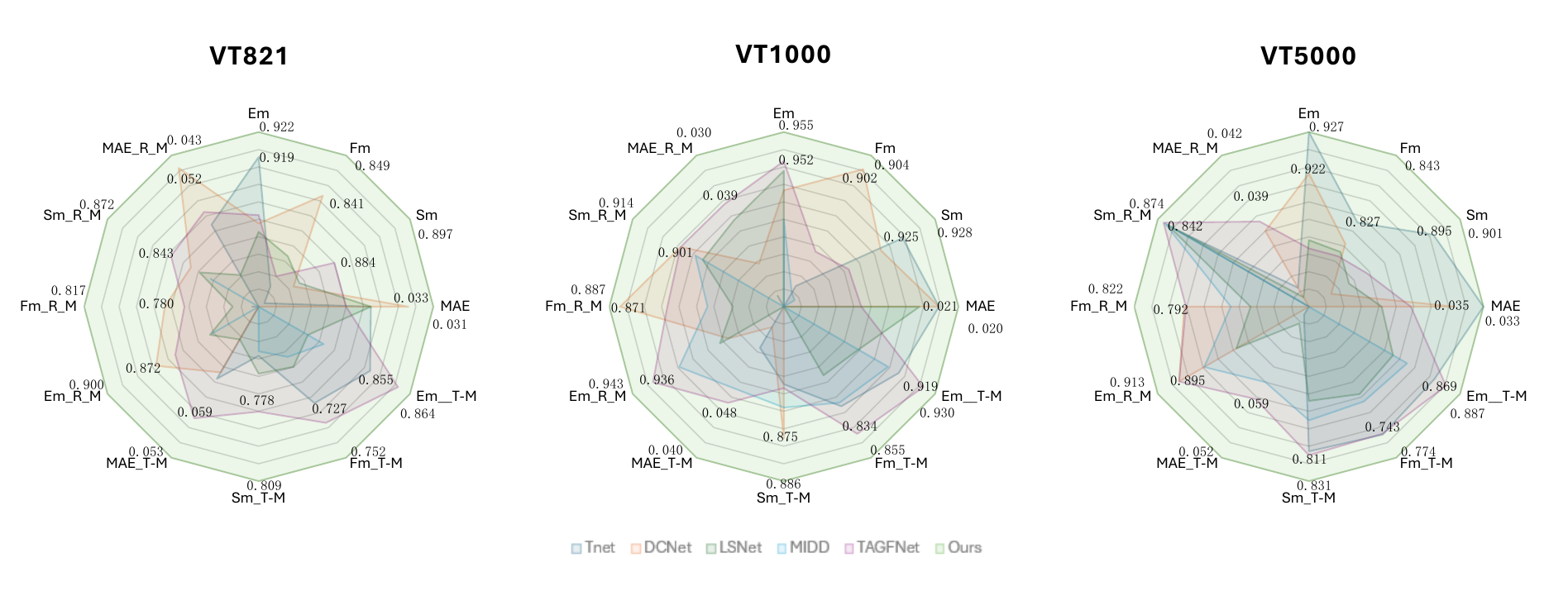}
\caption{Radar chart of the proposed model and some existing state-of-the-art methods in three datasets. R-M means the performance of the model under RGB-Missing. T-M means the performance of the model under Thermal-Missing. }
\label{radar chart}
\end{figure*}
As shown in Fig. \ref{radar chart}, we compared the best five models: MIDD \cite{tu2021multi}, DCNet \cite{tu2022weakly}, TNet \cite{cong2022does}, TAGFNet \cite{wang2023thermal}, LSNet \cite{zhou2023lsnet}, and our model to establish a radar chart. From the chart, it can be seen that CoLA achieved state-of-the-art performance across all metrics under the three conditions on all datasets.
\begin{table}[t]
\centering
\fontsize{7}{8}\selectfont
\caption{Compare with other image quality assessment methods for training stage \uppercase\expandafter{\romannumeral 1} . The Baseline is the network with a value of $\alpha$ set to 0.5. CLIP-IQA \cite{wang2023exploring} and CLIP-IQA$^+$ \cite{wang2023exploring} refer to the use of frozen CLIP and CLIP fine-tuned with the CoOp \cite{zhou2022learning} method, respectively.}
\begin{tabular}{llp{1.2cm}<{\centering}p{1.2cm}<{\centering}p{1.2cm}<{\centering}p{1.2cm}<{\centering}p{1.6cm}<{\centering}p{1.2cm}<{\centering}p{1.2cm}<{\centering}}
\toprule
& & Baseline&+BRISQUE \cite{mittal2012no} &+GIE \  \ \cite{cong2022does} & +CLIP-IQA \cite{wang2023exploring} & +CLIP-IQA$^+$\cite{wang2023exploring}& +LQA \\ \midrule
\multirow{4}{*}{VT821} & $S_\alpha$$\uparrow$ & .844 &.854&.870& .869 & .878 & \textbf{.888} \\
& $E_m$$\uparrow$ & .873 &.893&.910& .898 & .910 & \textbf{.915} \\
& $F_\beta$$\uparrow$ & .805 &.797&.812& .796 & .830  & \textbf{.839} \\
& MAE$\downarrow$ & .055 &.046&.044& .050 & .042  & \textbf{.038} \\
\midrule
\multirow{4}{*}{VT1000} & $S_\alpha$$\uparrow$ & .897 &.909&.920& .913 & .918 & \textbf{.924} \\
& $E_m$$\uparrow$ & .924 &.939&.944& .945 & .949  & \textbf{.955} \\
& $F_\beta$$\uparrow$ & .854 &.867&.880& .876 & .874 & \textbf{.904} \\
& MAE$\downarrow$ & .038 &.034&.030& .031 & .030  & \textbf{.024} \\ \midrule
\multicolumn{1}{c}{\multirow{4}{*}{VT5000}} & $S_\alpha$$\uparrow$ & .859&.864 &.866& .878 & .882  & \textbf{.887} \\
\multicolumn{1}{c}{} & $E_m$$\uparrow$ & .892&.901&.914 & .916 & .915 & \textbf{.923} \\
\multicolumn{1}{c}{} & $F_\beta$$\uparrow$ & .801 &.810&.819& .822 & .825& \textbf{.834} \\
\multicolumn{1}{c}{} & MAE$\downarrow$ & .052&.048 &.047& .042 & .040  & \textbf{.039} \\ 
 \bottomrule
\end{tabular}
\label{table_LQA Comparison}
\end{table}
\subsection{Ablation studies}
\textbf{Effectiveness of each component in our model. }
Table \ref{table_Ablation for all} summarizes how progressively incorporating our proposed modules improves model performance under modality-complete and modality-missing. 
Comparing (a) and (b) shows that adding the LQA module helps CoLA better fuse information from the two modalities to handle noisy images, thus achieving better results under modality-complete condition.
% Without LQA, the ability of pretrained model of the stage one is restricted compared with (b), so employing only Conditional Dropout in training stage \uppercase\expandafter{\romannumeral 2} bring weaker than (d).
Contrasting (c) and (d), the model trained without LQA in stage \uppercase\expandafter{\romannumeral 1} will be weaker than the model trained with LQA, which means employing only CD in stage \uppercase\expandafter{\romannumeral 2} results in weaker performance compared to (d).
% a b / a c :我们的模块带来提升
%ac 二阶段cd提升三个condition下的效果
%cd 没有LQA带来较弱的结果，性能下降
% b d / c d :使用一个会有局限 
% can enhance the model's capabilities under modality-missing, but it cannot improve the model's performance under modality-complete.
By incorporating both LQA and Conditional Dropout, the model significantly enhances its ability to extract valuable information from each individual modality. 
This fundamentally strengthens modal robustness, enabling CoLA to minimize performance loss when modalities are missing by better utilizing the available single modalities. \\\\
\textbf{The efficiency of LQA. }
Table \ref{table_LQA Comparison} compares other image quality assessment methods with LQA. In addition to the No-Reference\cite{mittal2012no} and Pre-trained quality assessment networks\cite{cong2022does}. 
We also compared the CLIP-IQA \cite{wang2023exploring}, which is based on CLIP\cite{radford2021learning}, to assess image quality. 
The results indicate that compared to using a fixed threshold of 0.5, employing image quality assessment methods such as BRISQUE\cite{mittal2012no}, GIE\cite{cong2022does}, CLIP-IQA \cite{wang2023exploring}, etc., can lead to some performance improvement. However, due to the lack of inherent learning capability and generalization to datasets, these methods have limited impact.
% Results demonstrate that incorporating the original CLIP does not improve model performance but rather impairs it while utilizing language-image models with prompt learning such as CoOp \cite{zhou2022learning} or CoCoOp \cite{zhou2022conditional} leads to moderate performance gains. Our proposed LQA module achieves superior performance compared to other approaches.

For better demonstrate the effectiveness of LQA, we extract all noisy images like \ref{motivation}(a) line 1 from original VT821 to create a new dataset VT821-noisy with 76 images. Experiments were conducted on this dataset using all compared methods, with results provided in the \textcolor{myPink}{supplementary materials}.
% As shown in Table \ref{table_LQA Comparison}, the improvement brought by the module is relatively small on the VT1000 dataset, which has less diverse scenes. In contrast, the improvement is more significant on the VT821 dataset with more diverse scenes, including varying times of day, lighting conditions, indoor/outdoor settings, and other environmental factors. Substantial gains are also attained with the LQA module on the VT5000 dataset.\\
\begin{table*}[t]
\caption{Ablation experiments of the Conditional Dropout module on the VT5000 dataset, ``Copy" denotes duplicating encoder, ``Freeze" denotes freezing modules except the copied encoder, and ``MD" denotes Modality Dropout. ``Z-Conv'' denotes Zero-Convolution.}
\centering
\fontsize{5}{8.5}\selectfont
\begin{tabular}{p{0.3cm}<{\centering}p{0.5cm}<{\centering}p{0.7cm}<{\centering}p{0.6cm}<{\centering}p{0.4cm}<{\centering}|*{3}{p{0.4cm}<{\centering}}p{0.7cm}<{\centering}|*{3}{p{0.4cm}<{\centering}}p{0.7cm}<{\centering}|*{3}{p{0.4cm}<{\centering}}p{0.7cm}<{\centering}|*{3}{p{0.4cm}<{\centering}}p{0.6cm}<{\centering}}
\toprule
\multirow{2}{*}{}& \multirow{2}{*}{Copy}& \multirow{2}{*}{Z-Conv} & \multirow{2}{*}{Freeze} & \multirow{2}{*}{MD} & \multicolumn{4}{c|}{Modality Complete} & \multicolumn{4}{c|}{Missing RGB} & \multicolumn{4}{c|}{Missing Thermal} & \multicolumn{4}{c}{Average} \\
 & && & & $S_\alpha$$\uparrow$ & $E_m$$\uparrow$ & $F_\beta$$\uparrow$ & MAE$\downarrow$ & $S_\alpha$$\uparrow$ & $E_m$$\uparrow$ & $F_\beta$$\uparrow$ & MAE$\downarrow$ & $S_\alpha$$\uparrow$ & $E_m$$\uparrow$ & $F_\beta$$\uparrow$ & MAE$\downarrow$ & $S_\alpha$$\uparrow$ & $E_m$$\uparrow$ & $F_\beta$$\uparrow$ & MAE$\downarrow$ \\
\midrule
(a) & & &  & &.887 &.923 &.834 &.039 &.828 &.876 &.754 &.059 &.849& .884&.789 &.061 &.855 &.894 &.792 &.053 \\
(b) &$\checkmark$  && &&.872 &.900&.813 &.049 &.815 &.860 &.735 &.066 &.851& .877&.788 &.062 &.857 &.894 &.786 &.054 \\
(c) &$\checkmark$  &$\checkmark$& &&.878 &.910&.819 &.046 &.820 &.863 &.733 &.062 &.847& .875&.785 &.066 &.857 &.894 &.786 &.054 \\
(d) &  && &$\checkmark$&.867 &.902&.805 &.051 &.839 &.880 &.759 &.057 &.865& .899&.795 &.054 &.857 &.894 &.786 &.054 \\
(e) &$\checkmark$  &$\checkmark$ & &$\checkmark$  &.872 &.913 &.822 &.046 &.834 &.875&.749 &.061 &.868 &.901 &.803 &.049 &.858 &.896 &.791 &.052 \\
(f) &$\checkmark$  &$\checkmark$ &$\checkmark$  & &.890 &.923 &.836 &.038 &.822 &.869 &.743 &.064 &.855 &.889 &.795 &.056 &.856 &.894 &.791 &.053 \\
(g) &$\checkmark$ & &$\checkmark$  &$\checkmark$ &.889 &.923 &.832 &.039 &.837 &.883 &.765 &.054 &.868 &.905 &.805 &.046 &.865 &.904 &.801 &.046 \\
\midrule
(h) &$\checkmark$  &$\checkmark$ &$\checkmark$  & $\checkmark$ &\textbf{.892} &\textbf{.927} &\textbf{.843} &\textbf{.037} &\textbf{.840} &\textbf{.887} &\textbf{.774} &\textbf{.052} &\textbf{.874} &\textbf{.913} &\textbf{.822} &\textbf{.042} &\textbf{.869} &\textbf{.909} &\textbf{.813} &\textbf{.044} \\
\bottomrule
\end{tabular}
\label{table_Ablation for CD}
\end{table*}
\\\\
\textbf{Ablation study for CD. }
Ablation experiments were conducted in this module to validate the effectiveness of the proposed Conditional Dropout. As shown in Table \ref{table_Ablation for CD}, observing (a) indicates that the original model lacks modality robustness, as evidenced by the significant performance reduction under modality-missing. Comparing (a) and (d) shows that simply applying Modality Dropout can improve model performance under modality-missing, but this leads to worse performance under modality-complete. Contrasting (d) and (e) demonstrates that adding the Copy operation can effectively reduce the performance decline caused by Modality Dropout under modality-complete. A comparison between (e) and (h) reveals that the absence of the freeze operation impacts the encoder's ability to effectively extract features, leading to diminished performance under modality-complete. Analyzing (f) and (h) suggests that employing Copy and Freeze without Modality Dropout does not enhance model robustness. 
% Finally, comparing (f) to the above configurations highlights that our proposed Conditional Dropout can concurrently enhance model robustness under modality-missing and maintain performance with modality-complete.
% Contrasting (b) and (c) demonstrates that adding the Copy operation can effectively reduce the performance decline caused by Modality Dropout under modality-complete. Further comparison between conditions (d) and (f) suggests that employing Copy and Freeze without Modality Dropout does not improve model robustness. Finally, comparing (f) to the above configurations highlights that our proposed Conditional Dropout can concurrently enhance model robustness under modality-missing and maintain performance with modality-complete. \\
\begin{figure*}[t]
\centering
\includegraphics[width=0.9\linewidth,keepaspectratio]{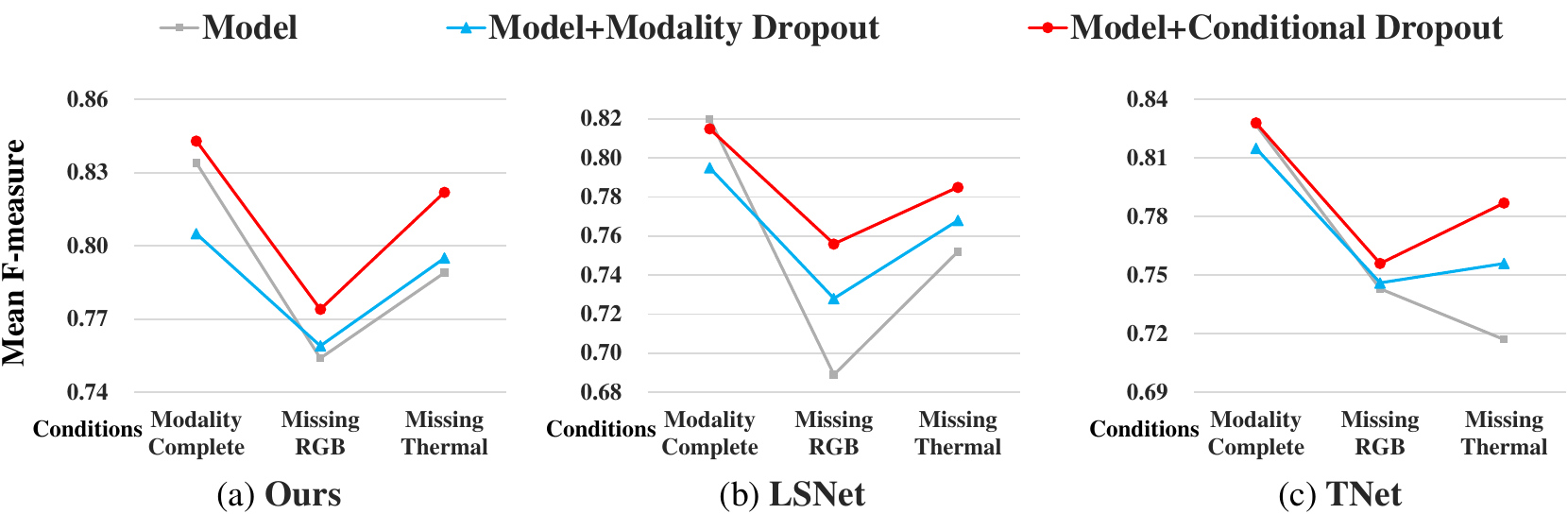}
\caption{Comparison of Models Trained with Conditional Dropout, Modality Dropout, and Original Model in VT5000 dataset.}
\label{fig_plug-in}
\end{figure*}
\\\\
\textbf{Conditional Dropout as a plug-in. }
We conducted generalization experiments to validate the effectiveness of Conditional Dropout. The results in Fig. \ref{fig_plug-in} show that LSNet \cite{zhou2023lsnet} and T-Net \cite{cong2022does} augmented with Conditional Dropout achieves significantly improved robustness under modality-missing while the performance under modality-complete remains almost unaffected. These experiments validate the generalization of Conditional Dropout in advancing the robustness of dual-modal methods.
% To further demonstrate the efficacy of Conditional Dropout, additional experiments were performed by applying it to other datasets.
\\\\
\textbf{Extended experiments under modality-complete with CD. }
As shown in Table \ref{table_Validation for modality complete}, the model demonstrates improved performance under modality-complete after applying Conditional Dropout, especially on the VT821 dataset. Comparing (a) with (b) and (d) reveals that simply extending training epochs or copying parameters fails to enhance performance under modality-complete. Contrasting (c) to (a) indicates that directly utilizing Modality Dropout damages performance under complete modalities. Comparing (c) and (e), it can be observed that Conditional Dropout, as opposed to Modality Dropout, can significantly enhance the model's performance in modality-complete.

Overall, simply increasing model parameters (copying encoders), training epochs, or using Modality Dropout does not improve the model's performance under modality-complete. Conditional Dropout enhances the model's ability to extract information from individual modality, allowing the model to maintain or even improve its performance under modality-complete.

\begin{table*}[t]
\caption{Extended experiments under modality-complete. 
``AT" denotes additional training for 60 epochs.}
\centering
\fontsize{6.5}{9.5}\selectfont
\begin{tabular}{p{0.3cm}p{0.4cm}p{0.6cm}p{0.8cm}p{0.7cm}p{0.6cm}|*{3}{p{0.5cm}}p{0.7cm}|*{3}{p{0.5cm}}p{0.7cm}|*{3}{p{0.5cm}}p{0.7cm}}
\toprule
\multirow{2}{*}{} & \multirow{2}{*}{AT}& \multirow{2}{*}{Copy}& \multirow{2}{*}{Z-Conv} & \multirow{2}{*}{Freeze} & \multirow{2}{*}{MD} & \multicolumn{4}{c|}{VT821} & \multicolumn{4}{c|}{VT1000} & \multicolumn{4}{c}{VT5000}  \\
 & & && && $S_\alpha$$\uparrow$ & $E_m$$\uparrow$ & $F_\beta$$\uparrow$ & MAE$\downarrow$ & $S_\alpha$$\uparrow$ & $E_m$$\uparrow$ & $F_\beta$$\uparrow$ & MAE$\downarrow$ & $S_\alpha$$\uparrow$ & $E_m$$\uparrow$ & $F_\beta$$\uparrow$ & MAE$\downarrow$  \\
\midrule
(a) & & && & &.888 &.915 &.839 &.038 &.924 &.955 &.893 &.024 &.887 &.923 &.834 &.039  \\
(b) &$\checkmark$& && &  &.870 &.881 &.806 &.051 & .915&.944 &.879 &.031 &.871 &.910 &.810 &.046  \\
(c) &$\checkmark$& & &  &$\checkmark$ &.851 &.872 &.788 &.055 &.910 &.935 &.870 &.034 &.867 &.902 &.805 &.051  \\
(d) &$\checkmark$&$\checkmark$  &$\checkmark$&  &&.878 &.905 &.832 &.043 &.922 &.948 &.879 &.029 &.878 &.910 &.819 &.042   \\
% (e) &$\checkmark$&$\checkmark$ & &$\checkmark$  &$\checkmark$ &.891 &.917 &.838 &.040 &.928 &.955 &.890 &.024 &.889 &.923 &.832 &.039  \\
\midrule
(e) &$\checkmark$&$\checkmark$  &$\checkmark$ &$\checkmark$  & $\checkmark$ &\textbf{.897} &\textbf{.922} &\textbf{.849} 
&\textbf{.031} &\textbf{.928} &\textbf{.955} &\textbf{.904} &\textbf{.024} &\textbf{.892} &\textbf{.927} &\textbf{.843} 
&\textbf{.037}  \\
\bottomrule
\end{tabular}
\label{table_Validation for modality complete}
\end{table*}
\section{Conclusion}
In this paper, we proposed a robust dual-modal SOD method to enhance model performance in noisy and modality-missing environments. We adeptly adjust reweighted image contributions by the proposed LQA, which is crucial for noise handling. This method bolsters noise robustness across diverse environments. We further introduce a Conditional Dropout training scheme, effective in processing missing modalities, thus improving model performance with incomplete and complete inputs. Our approach sets a new evaluation in dual-modal SOD, prompting discussions on handling robustness in the SOD field.

\clearpage  % TODO REVIEW/FINAL: This \clearpage needs to be removed from both review and camera-ready versions.

\bibliographystyle{splncs04}
\bibliography{main}

\begin{thebibliography}{10}
\providecommand{\url}[1]{\texttt{#1}}
\providecommand{\urlprefix}{URL }
\providecommand{\doi}[1]{https://doi.org/#1}

\bibitem{achanta2009frequency}
Achanta, R., Hemami, S., Estrada, F., Susstrunk, S.: Frequency-tuned salient region detection. In: 2009 IEEE conference on computer vision and pattern recognition. pp. 1597--1604. IEEE (2009)

\bibitem{borji2015salient}
Borji, A., Cheng, M.M., Jiang, H., Li, J.: Salient object detection: A benchmark. IEEE transactions on image processing  \textbf{24}(12),  5706--5722 (2015)

\bibitem{cai2018deep}
Cai, L., Wang, Z., Gao, H., Shen, D., Ji, S.: Deep adversarial learning for multi-modality missing data completion. In: Proceedings of the 24th ACM SIGKDD international conference on knowledge discovery \& data mining. pp. 1158--1166 (2018)

\bibitem{chen2021depth}
Chen, C., Wei, J., Peng, C., Qin, H.: Depth-quality-aware salient object detection. IEEE Transactions on Image Processing  \textbf{30},  2350--2363 (2021)

\bibitem{chen2022saliency}
Chen, T., Yao, Y., Zhang, L., Wang, Q., Xie, G., Shen, F.: Saliency guided inter-and intra-class relation constraints for weakly supervised semantic segmentation. IEEE Transactions on Multimedia  (2022)

\bibitem{cheng2022depth}
Cheng, X., Zheng, X., Pei, J., Tang, H., Lyu, Z., Chen, C.: Depth-induced gap-reducing network for rgb-d salient object detection: an interaction, guidance and refinement approach. IEEE Transactions on Multimedia  (2022)

\bibitem{cheng2014depth}
Cheng, Y., Fu, H., Wei, X., Xiao, J., Cao, X.: Depth enhanced saliency detection method. In: Proceedings of international conference on internet multimedia computing and service. pp. 23--27 (2014)

\bibitem{cong2022cir}
Cong, R., Lin, Q., Zhang, C., Li, C., Cao, X., Huang, Q., Zhao, Y.: Cir-net: Cross-modality interaction and refinement for rgb-d salient object detection. IEEE Transactions on Image Processing  \textbf{31},  6800--6815 (2022)

\bibitem{cong2022does}
Cong, R., Zhang, K., Zhang, C., Zheng, F., Zhao, Y., Huang, Q., Kwong, S.: Does thermal really always matter for rgb-t salient object detection? IEEE Transactions on Multimedia  (2022)

\bibitem{cui2022survival}
Cui, C., Liu, H., Liu, Q., Deng, R., Asad, Z., Wang, Y., Zhao, S., Yang, H., Landman, B.A., Huo, Y.: Survival prediction of brain cancer with incomplete radiology, pathology, genomic, and demographic data. In: International Conference on Medical Image Computing and Computer-Assisted Intervention. pp. 626--635. Springer (2022)

\bibitem{deng2009imagenet}
Deng, J., Dong, W., Socher, R., Li, L.J., Li, K., Fei-Fei, L.: Imagenet: A large-scale hierarchical image database. In: 2009 IEEE conference on computer vision and pattern recognition. pp. 248--255. Ieee (2009)

\bibitem{ding2021rfnet}
Ding, Y., Yu, X., Yang, Y.: Rfnet: Region-aware fusion network for incomplete multi-modal brain tumor segmentation. In: Proceedings of the IEEE/CVF international conference on computer vision. pp. 3975--3984 (2021)

\bibitem{fan2017structure}
Fan, D.P., Cheng, M.M., Liu, Y., Li, T., Borji, A.: Structure-measure: A new way to evaluate foreground maps. In: Proceedings of the IEEE international conference on computer vision. pp. 4548--4557 (2017)

\bibitem{fan2018enhanced}
Fan, D.P., Gong, C., Cao, Y., Ren, B., Cheng, M.M., Borji, A.: Enhanced-alignment measure for binary foreground map evaluation. arXiv preprint arXiv:1805.10421  (2018)

\bibitem{fan2020rethinking}
Fan, D.P., Lin, Z., Zhang, Z., Zhu, M., Cheng, M.M.: Rethinking rgb-d salient object detection: Models, data sets, and large-scale benchmarks. IEEE Transactions on neural networks and learning systems  \textbf{32}(5),  2075--2089 (2020)

\bibitem{gurkan2021tdiot}
Gurkan, F., Cerkezi, L., Cirakman, O., Gunsel, B.: Tdiot: Target-driven inference for deep video object tracking. IEEE Transactions on Image Processing  \textbf{30},  7938--7951 (2021)

\bibitem{he2016deep}
He, K., Zhang, X., Ren, S., Sun, J.: Deep residual learning for image recognition. In: Proceedings of the IEEE conference on computer vision and pattern recognition. pp. 770--778 (2016)

\bibitem{huo2021efficient}
Huo, F., Zhu, X., Zhang, L., Liu, Q., Shu, Y.: Efficient context-guided stacked refinement network for rgb-t salient object detection. IEEE Transactions on Circuits and Systems for Video Technology  \textbf{32}(5),  3111--3124 (2021)

\bibitem{huo2022real}
Huo, F., Zhu, X., Zhang, Q., Liu, Z., Yu, W.: Real-time one-stream semantic-guided refinement network for rgb-thermal salient object detection. IEEE Transactions on Instrumentation and Measurement  \textbf{71},  1--12 (2022)

\bibitem{jia2021scaling}
Jia, C., Yang, Y., Xia, Y., Chen, Y.T., Parekh, Z., Pham, H., Le, Q., Sung, Y.H., Li, Z., Duerig, T.: Scaling up visual and vision-language representation learning with noisy text supervision. In: International conference on machine learning. pp. 4904--4916. PMLR (2021)

\bibitem{john2023multimodal}
John, V., Kawanishi, Y.: Multimodal cascaded framework with metric learning robust to missing modalities for person classification. In: Proceedings of the 14th Conference on ACM Multimedia Systems. pp. 257--265 (2023)

\bibitem{ju2014depth}
Ju, R., Ge, L., Geng, W., Ren, T., Wu, G.: Depth saliency based on anisotropic center-surround difference. In: 2014 IEEE international conference on image processing (ICIP). pp. 1115--1119. IEEE (2014)

\bibitem{jue2019integrating}
Jue, J., Jason, H., Neelam, T., Andreas, R., Sean, B.L., Joseph, D.O., Harini, V.: Integrating cross-modality hallucinated mri with ct to aid mediastinal lung tumor segmentation. In: Medical Image Computing and Computer Assisted Intervention--MICCAI 2019: 22nd International Conference, Shenzhen, China, October 13--17, 2019, Proceedings, Part VI 22. pp. 221--229. Springer (2019)

\bibitem{kong2021spatiotemporal}
Kong, Y., Wang, Y., Li, A.: Spatiotemporal saliency representation learning for video action recognition. IEEE Transactions on Multimedia  \textbf{24},  1515--1528 (2021)

\bibitem{lee2022spsn}
Lee, M., Park, C., Cho, S., Lee, S.: Spsn: Superpixel prototype sampling network for rgb-d salient object detection. In: European Conference on Computer Vision. pp. 630--647. Springer (2022)

\bibitem{lee2023saliency}
Lee, M., Lee, S., Lee, J., Shim, H.: Saliency as pseudo-pixel supervision for weakly and semi-supervised semantic segmentation. IEEE Transactions on Pattern Analysis and Machine Intelligence  (2023)

\bibitem{liang2023crowdclip}
Liang, D., Xie, J., Zou, Z., Ye, X., Xu, W., Bai, X.: Crowdclip: Unsupervised crowd counting via vision-language model. In: Proceedings of the IEEE/CVF Conference on Computer Vision and Pattern Recognition. pp. 2893--2903 (2023)

\bibitem{liu2023sgfusion}
Liu, J., Dian, R., Li, S., Liu, H.: Sgfusion: A saliency guided deep-learning framework for pixel-level image fusion. Information Fusion  \textbf{91},  205--214 (2023)

\bibitem{liu2023vlpd}
Liu, M., Jiang, J., Zhu, C., Yin, X.C.: Vlpd: Context-aware pedestrian detection via vision-language semantic self-supervision. In: Proceedings of the IEEE/CVF Conference on Computer Vision and Pattern Recognition. pp. 6662--6671 (2023)

\bibitem{liu2024vst++}
Liu, N., Luo, Z., Zhang, N., Han, J.: Vst++: Efficient and stronger visual saliency transformer. IEEE Transactions on Pattern Analysis and Machine Intelligence  (2024)

\bibitem{liu2020learning}
Liu, N., Zhang, N., Han, J.: Learning selective self-mutual attention for rgb-d saliency detection. In: Proceedings of the IEEE/CVF conference on computer vision and pattern recognition. pp. 13756--13765 (2020)

\bibitem{liu2021learning}
Liu, N., Zhang, N., Shao, L., Han, J.: Learning selective mutual attention and contrast for rgb-d saliency detection. IEEE Transactions on Pattern Analysis and Machine Intelligence  \textbf{44}(12),  9026--9042 (2021)

\bibitem{liu2021visual}
Liu, N., Zhang, N., Wan, K., Shao, L., Han, J.: Visual saliency transformer. In: Proceedings of the IEEE/CVF international conference on computer vision. pp. 4722--4732 (2021)

\bibitem{liu2021swin}
Liu, Z., Lin, Y., Cao, Y., Hu, H., Wei, Y., Zhang, Z., Lin, S., Guo, B.: Swin transformer: Hierarchical vision transformer using shifted windows. In: Proceedings of the IEEE/CVF international conference on computer vision. pp. 10012--10022 (2021)

\bibitem{ma2022multimodal}
Ma, M., Ren, J., Zhao, L., Testuggine, D., Peng, X.: Are multimodal transformers robust to missing modality? In: Proceedings of the IEEE/CVF Conference on Computer Vision and Pattern Recognition. pp. 18177--18186 (2022)

\bibitem{mittal2012no}
Mittal, A., Moorthy, A.K., Bovik, A.C.: No-reference image quality assessment in the spatial domain. IEEE Transactions on image processing  \textbf{21}(12),  4695--4708 (2012)

\bibitem{ning2023hoiclip}
Ning, S., Qiu, L., Liu, Y., He, X.: Hoiclip: Efficient knowledge transfer for hoi detection with vision-language models. In: Proceedings of the IEEE/CVF Conference on Computer Vision and Pattern Recognition. pp. 23507--23517 (2023)

\bibitem{pan2020spatially}
Pan, Y., Liu, M., Lian, C., Xia, Y., Shen, D.: Spatially-constrained fisher representation for brain disease identification with incomplete multi-modal neuroimages. IEEE transactions on medical imaging  \textbf{39}(9),  2965--2975 (2020)

\bibitem{pang2023caver}
Pang, Y., Zhao, X., Zhang, L., Lu, H.: Caver: Cross-modal view-mixed transformer for bi-modal salient object detection. IEEE Transactions on Image Processing  \textbf{32},  892--904 (2023)

\bibitem{peng2014rgbd}
Peng, H., Li, B., Xiong, W., Hu, W., Ji, R.: Rgbd salient object detection: A benchmark and algorithms. In: Computer Vision--ECCV 2014: 13th European Conference, Zurich, Switzerland, September 6-12, 2014, Proceedings, Part III 13. pp. 92--109. Springer (2014)

\bibitem{radford2021learning}
Radford, A., Kim, J.W., Hallacy, C., Ramesh, A., Goh, G., Agarwal, S., Sastry, G., Askell, A., Mishkin, P., Clark, J., et~al.: Learning transferable visual models from natural language supervision. In: International conference on machine learning. pp. 8748--8763. PMLR (2021)

\bibitem{ramesh2021zero}
Ramesh, A., Pavlov, M., Goh, G., Gray, S., Voss, C., Radford, A., Chen, M., Sutskever, I.: Zero-shot text-to-image generation. In: International Conference on Machine Learning. pp. 8821--8831. PMLR (2021)

\bibitem{tu2021multi}
Tu, Z., Li, Z., Li, C., Lang, Y., Tang, J.: Multi-interactive dual-decoder for rgb-thermal salient object detection. IEEE Transactions on Image Processing  \textbf{30},  5678--5691 (2021)

\bibitem{tu2022weakly}
Tu, Z., Li, Z., Li, C., Tang, J.: Weakly alignment-free rgbt salient object detection with deep correlation network. IEEE Transactions on Image Processing  \textbf{31},  3752--3764 (2022)

\bibitem{tu2022rgbt}
Tu, Z., Ma, Y., Li, Z., Li, C., Xu, J., Liu, Y.: Rgbt salient object detection: A large-scale dataset and benchmark. IEEE Transactions on Multimedia  (2022)

\bibitem{tu2020rgbt}
Tu, Z., Xia, T., Li, C., Wang, X., Ma, Y., Tang, J.: Rgb-t image saliency detection via collaborative graph learning. IEEE Transactions on Multimedia  \textbf{22}(1),  160--173 (2020). \doi{10.1109/TMM.2019.2924578}

\bibitem{wang2018rgb}
Wang, G., Li, C., Ma, Y., Zheng, A., Tang, J., Luo, B.: Rgb-t saliency detection benchmark: Dataset, baselines, analysis and a novel approach. In: Image and Graphics Technologies and Applications: 13th Conference on Image and Graphics Technologies and Applications, IGTA 2018, Beijing, China, April 8--10, 2018, Revised Selected Papers 13. pp. 359--369. Springer (2018)

\bibitem{wang2023thermal}
Wang, H., Song, K., Huang, L., Wen, H., Yan, Y.: Thermal images-aware guided early fusion network for cross-illumination rgb-t salient object detection. Engineering Applications of Artificial Intelligence  \textbf{118},  105640 (2023)

\bibitem{wang2023exploring}
Wang, J., Chan, K.C., Loy, C.C.: Exploring clip for assessing the look and feel of images. In: Proceedings of the AAAI Conference on Artificial Intelligence. vol.~37, pp. 2555--2563 (2023)

\bibitem{wang2021cgfnet}
Wang, J., Song, K., Bao, Y., Huang, L., Yan, Y.: Cgfnet: Cross-guided fusion network for rgb-t salient object detection. IEEE Transactions on Circuits and Systems for Video Technology  \textbf{32}(5),  2949--2961 (2021)

\bibitem{wasim2023vita}
Wasim, S.T., Naseer, M., Khan, S., Khan, F.S., Shah, M.: Vita-clip: Video and text adaptive clip via multimodal prompting. In: Proceedings of the IEEE/CVF Conference on Computer Vision and Pattern Recognition. pp. 23034--23044 (2023)

\bibitem{wei2023mmanet}
Wei, S., Luo, C., Luo, Y.: Mmanet: Margin-aware distillation and modality-aware regularization for incomplete multimodal learning. In: Proceedings of the IEEE/CVF Conference on Computer Vision and Pattern Recognition. pp. 20039--20049 (2023)

\bibitem{wu2023hidanet}
Wu, Z., Allibert, G., Meriaudeau, F., Ma, C., Demonceaux, C.: Hidanet: Rgb-d salient object detection via hierarchical depth awareness. IEEE Transactions on Image Processing  \textbf{32},  2160--2173 (2023)

\bibitem{yu2023task}
Yu, T., Lu, Z., Jin, X., Chen, Z., Wang, X.: Task residual for tuning vision-language models. In: Proceedings of the IEEE/CVF Conference on Computer Vision and Pattern Recognition. pp. 10899--10909 (2023)

\bibitem{yu2023turning}
Yu, W., Liu, Y., Hua, W., Jiang, D., Ren, B., Bai, X.: Turning a clip model into a scene text detector. In: Proceedings of the IEEE/CVF Conference on Computer Vision and Pattern Recognition. pp. 6978--6988 (2023)

\bibitem{zhang2023adding}
Zhang, L., Rao, A., Agrawala, M.: Adding conditional control to text-to-image diffusion models. In: Proceedings of the IEEE/CVF International Conference on Computer Vision. pp. 3836--3847 (2023)

\bibitem{zhang2022c}
Zhang, M., Yao, S., Hu, B., Piao, Y., Ji, W.: C$^{2}$dfnet: Criss-cross dynamic filter network for rgb-d salient object detection. IEEE Transactions on Multimedia  (2022)

\bibitem{zhang2022can}
Zhang, R., Zeng, Z., Guo, Z., Li, Y.: Can language understand depth? In: Proceedings of the 30th ACM International Conference on Multimedia. pp. 6868--6874 (2022)

\bibitem{zhou2022conditional}
Zhou, K., Yang, J., Loy, C.C., Liu, Z.: Conditional prompt learning for vision-language models. In: Proceedings of the IEEE/CVF Conference on Computer Vision and Pattern Recognition. pp. 16816--16825 (2022)

\bibitem{zhou2022learning}
Zhou, K., Yang, J., Loy, C.C., Liu, Z.: Learning to prompt for vision-language models. International Journal of Computer Vision  \textbf{130}(9),  2337--2348 (2022)

\bibitem{zhou2023lsnet}
Zhou, W., Zhu, Y., Lei, J., Yang, R., Yu, L.: Lsnet: Lightweight spatial boosting network for detecting salient objects in rgb-thermal images. IEEE Transactions on Image Processing  \textbf{32},  1329--1340 (2023)

\bibitem{zhou2021saliency}
Zhou, Z., Pei, W., Li, X., Wang, H., Zheng, F., He, Z.: Saliency-associated object tracking. In: Proceedings of the IEEE/CVF international conference on computer vision. pp. 9866--9875 (2021)

\end{thebibliography}

\appendix
\title{CoLA: Conditional Dropout and Language-driven Robust Dual-modal Salient Object Detection \\ <Supplementary Material>} 

\titlerunning{CoLA: Dual-modal SOD}

\author{Shuang Hao\inst{1}\textsuperscript{$\star$}\orcidlink{0009-0007-4621-8485} \and
Chunlin Zhong\inst{1}\textsuperscript{$\star$}\orcidlink{0009-0006-7070-3504} \and
He Tang\inst{1}\textsuperscript{\Letter}\orcidlink{0000-0002-8454-1407}}

\authorrunning{S. Hao et al.}

\institute{School of Software Engineering, Huazhong University of Science and Technology, Wuhan, China\\
\email{\{shuanghao, clzhong, hetang\}@hust.edu.cn}
\let\thefootnote\relax\footnotetext{\textsuperscript{$\star$} Equal contribution \quad \textsuperscript{\Letter} Corresponding author}
}

\maketitle
% \newcommand{\solidcircle}{%
%     \begin{tikzpicture}
%         \fill[black](0,0) circle (0.1cm);
%     \end{tikzpicture}%
% }
% \newcommand{\emptycircle}{%
%     \begin{tikzpicture}
%         \draw[black](0,0) circle (0.1cm);
%     \end{tikzpicture}%
% }
% \definecolor{myRed}{RGB}{219, 68, 55}
% \definecolor{myGreen}{RGB}{15, 157, 88}
% \definecolor{myBlue}{RGB}{66, 133, 244}

Our supplementary material is divided into three parts, \cref{sec:More Experiments of RGB-T inputs}, \cref{sec:More Experiments of RGB-D inputs} and \cref{sec:limitation}, focusing on RGB-T inputs, RGB-D inputs and limitation respectively, with the following organization.

Experiments on changing the backbone are in \cref{sec:Experiments on Various Backbones}; in \cref{sec:Quantitative Evaluation of LQA}, We have provided some examples to demonstrate the reliability of the $\alpha$ generated by LQA. In \cref{sec:More implementations of Conditional Dropout as a Plug-in}, we appended Conditional Dropout to VT1000 and VT821 datasets to further prove the generalizability of our method. For fairness in input methods, in \cref{sec:Comparative Analysis Under Modality Dropout}, we retrained all comparison methods using modality dropout and compared them with our method CoLA; in \cref{sec:Average and Average dropout}, we introduced the calculation methods and meanings of Average and Average Dropout mentioned in Table \textcolor{red}{1}. \cref{sec:Experiments on Noisy Inputs} describes related experiments for dividing the VT821-noisy subset. \cref{sec:Quantitative Evaluation} is a comparison of CoLA with the SOTA methods in the RGB-D field, and \cref{sec:Qualitative Evaluation} involves a qualitative comparison with the SOTA methods. \cref{sec:limitation} discusses the limitations of our method and presents examples where it did not perform well.

\section{More Experiments of RGB-T inputs}
\label{sec:More Experiments of RGB-T inputs}
% \begin{figure*}[t]
% \centering
% \includegraphics[width=1\linewidth,keepaspectratio]{img/plug_vt1000.pdf}
% \caption{Comparison of Models Trained with Conditional Dropout, Modality Dropout, and Original Model in VT1000 dataset.}
% \label{supp_-CD_VT1000}
% \end{figure*}
% \begin{figure*}[t]
% \centering
% \includegraphics[width=1\linewidth,keepaspectratio]{img/plug_vt821.pdf}
% \caption{Comparison of Models Trained with Conditional Dropout, Modality Dropout, and Original Model in VT821 dataset.}
% \label{supp_-CD_VT821}
% \end{figure*}

\subsection{Quantitative Evaluation of LQA}
\label{sec:Quantitative Evaluation of LQA}
\begin{figure*}[ht!]
\centering
\includegraphics[width=1\linewidth,keepaspectratio]{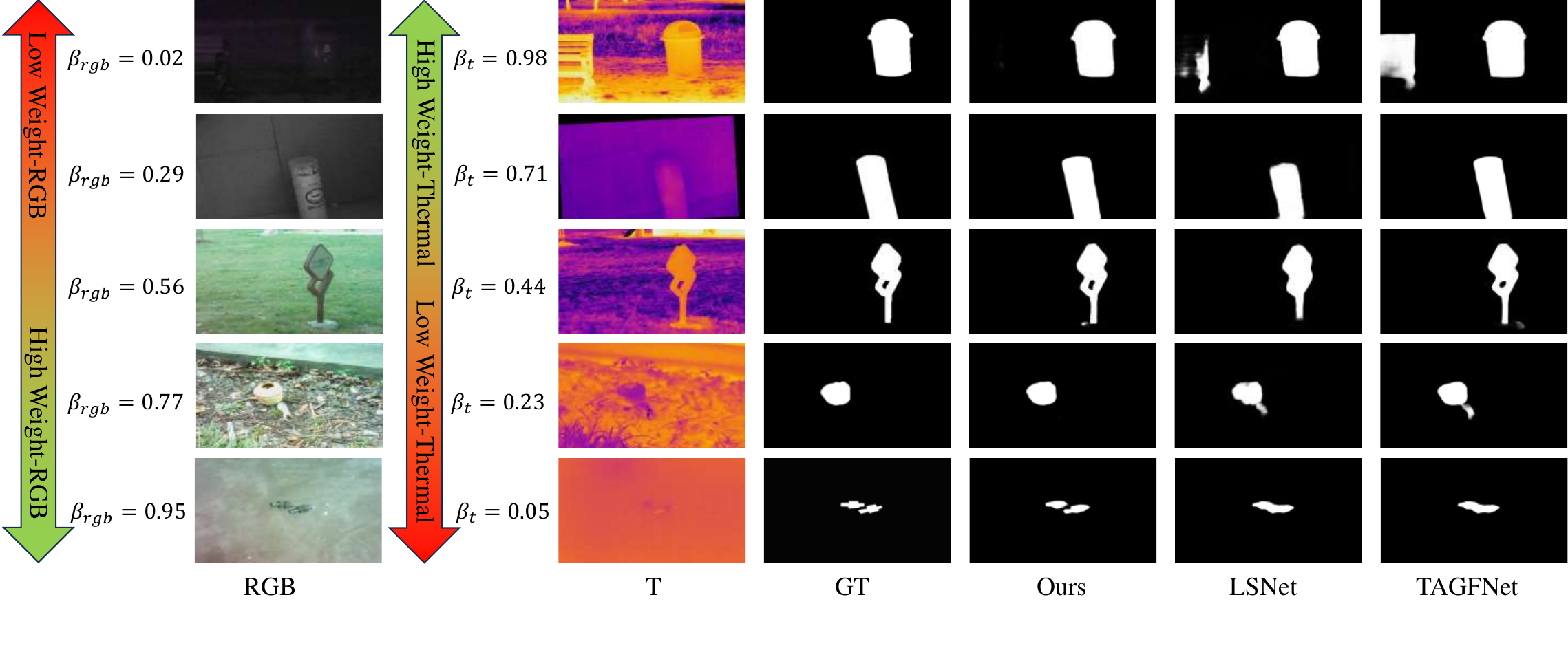}
\caption{Different cases of our LQA, $\beta_{rgb} = \frac{\alpha^{rgb}}{\alpha^{rgb}+\alpha^{t}}$,$\beta_{t} = \frac{\alpha^{t}}{\alpha^{rgb}+\alpha^{t}}$.}
\label{lqa}
\end{figure*}

As shown in the Fig \ref{lqa}, we present the quality assessment of LQA for different images. In the images of the first and last rows, the quality of the RGB or Thermal images is extremely low, rendering them unable to provide effective information. Correspondingly, the $\beta_{rgb}$ also approach 0 or 1, indicating that LQA can effectively handle these two extreme quality image scenarios. In the second to fourth lines, it can be observed that the quality of RGB images improves in comparison to thermal images, and the corresponding alpha values also increase. In conclusion, it has been demonstrated that LQA is not only effective in assessing the contribution of noisy images but also performs well on general images.

\subsection{More Implementations of Conditional Dropout as a Plug-in}
\label{sec:More implementations of Conditional Dropout as a Plug-in}
\begin{figure}[t!]
\centering
\begin{minipage}{1\linewidth}
		\centering
		\includegraphics[width=0.9\linewidth]{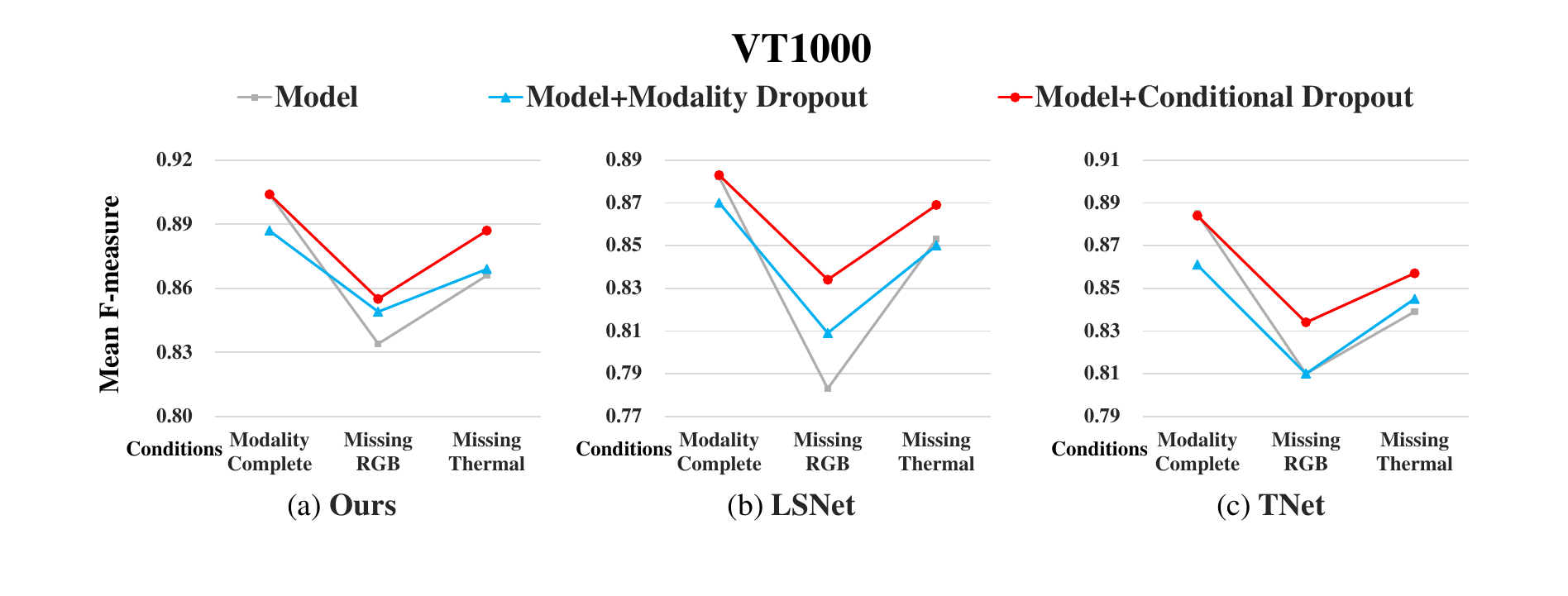}
		% \caption{failure cases of our model}
		\label{supp_-CD_VT1000}%文中引用该图片代号
	\end{minipage}
	\begin{minipage}{1\linewidth}
		\centering
		\includegraphics[width=0.9\linewidth]{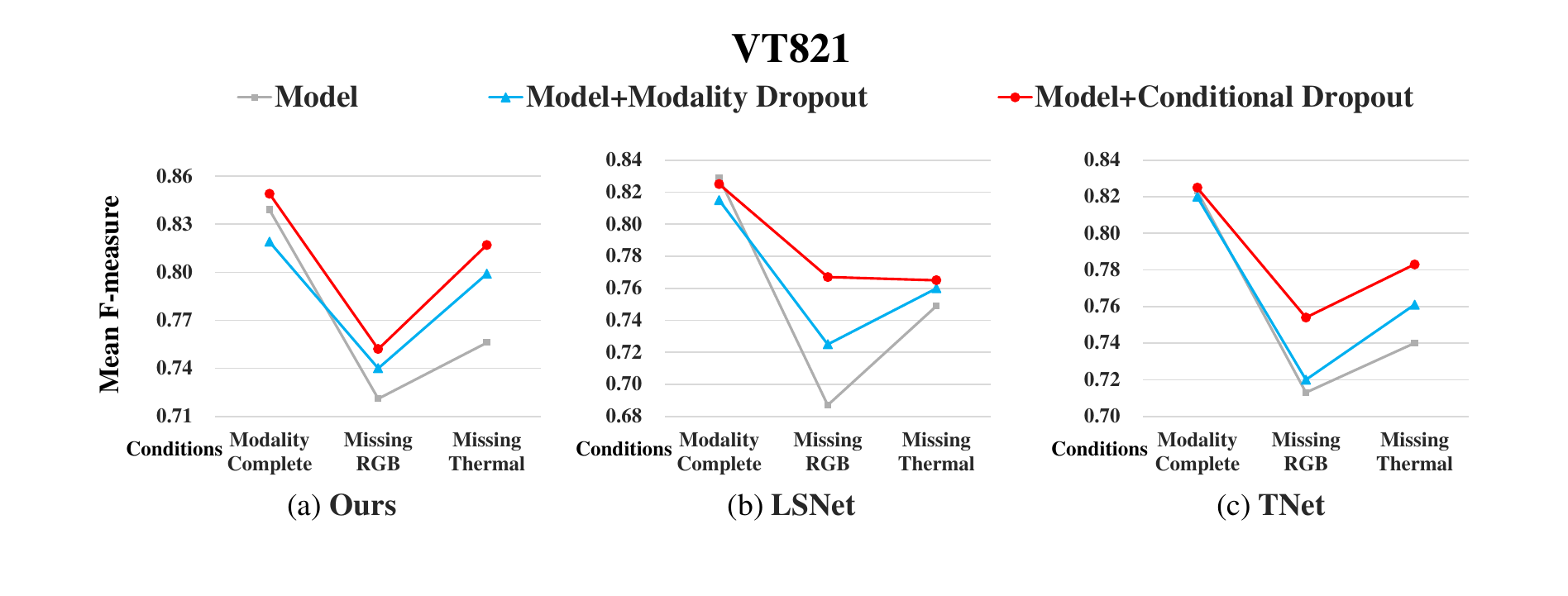}
		\caption{Comparison of Models Trained with Conditional Dropout, Modality Dropout, and Original Model in VT1000 and VT821 datasets.}
		\label{supp_-CD_VT821}
	\end{minipage}
\end{figure}

We also applied Conditional Dropout to VT1000 and VT821 to validate its effectiveness, and the experimental results are shown in \cref{supp_-CD_VT821}. We found that with Conditional Dropout, model's performance remains almost unchanged when all modalitiy-complete. Furthermore, its performance in two modality-missing conditions is significantly better than the original model and model using Modality Dropout. This effect is consistent across three different datasets, further demonstrating that Conditional Dropout enhances the model's robustness and reduces performance degradation in modality-missing conditions.

\subsection{Comparative Analysis Under Modality Dropout}
\label{sec:Comparative Analysis Under Modality Dropout}
Using the modality dropout training method may bring certain improvements when modalities are missing, but it can harm the performance under complete modality. Our proposed Conditional Dropout can simultaneously enhance the model's performance in both missing and complete modality scenarios. To further validate the effectiveness of our CoLA in RGB-T Salient Object Detection (SOD) under both modality-complete and modality-missing scenarios, we compared it with all the methods mentioned in the Sec. \textcolor{red}{4} under modality dropout training. The compared methods include ADF \cite{tu2022rgbt}, MIDD \cite{tu2021multi}, CSRNet \cite{huo2021efficient}, DCNet \cite{tu2022weakly}, TNet \cite{cong2022does}, TAGFNet \cite{wang2023thermal} and LSNet \cite{zhou2023lsnet}.

The quantitative comparison results are shown in Table \ref{supp_comparison with others in MD}. Compared with Table \textcolor{Red}{1}, comparative methods are mainly divided into two categories. One is that after using Modality Dropout, it can enhance its performance when modality is missing, but it will impair the model's performance when the modality is complete such as DCNet \cite{tu2022weakly} and LSNet \cite{zhou2023lsnet}. The other methods like TAGFNet \cite{wang2023thermal} is that using Modality Dropout will impair performance in both situations. Our method, even when Modality Dropout is used in other methods, has achieved the best results in both cases of modality-missing.
\begin{table}[t]
\centering
\caption{Comparison Results of Different Methods on the VT821-noisy Dataset. }
\fontsize{8.5}{11}\selectfont
\begin{tabular}{p{2.5cm}<{\centering}|*{3}{p{1cm}}p{1.2cm}}
\toprule
\multicolumn{1}{c|}{Methods}& \multicolumn{4}{c}{VT821-noisy}\\
& $S_\alpha$$\uparrow$ & $E_m$$\uparrow$ & $F_\beta$$\uparrow$ & MAE$\downarrow$\\
\midrule
% \multicolumn{17}{l}{\textbf{Performance of RGB-D Models under Modality-Complete}}\\
CSRNet \cite{huo2021efficient}  & 0.736             & 0.765         & 0.632         & 0.110  \\
ADF \cite{tu2022rgbt}     & 0.579             & 0.564         & 0.356         & 0.147  \\
TNet \cite{cong2022does}   & 0.807             & 0.826         & 0.729         & 0.055  \\
DCNet \cite{tu2022weakly}   & 0.818             & 0.853         & 0.752         & 0.056  \\
LSNet \cite{zhou2023lsnet}   & 0.804             & 0.823         & 0.732         & 0.052  \\
MIDD \cite{tu2021multi}   & 0.812             & 0.814         & 0.705         & 0.073  \\
TAGFNet \cite{wang2023thermal} & 0.772             & 0.794         & 0.651         & 0.074  \\
Ours    & \textcolor{myRed}{\textbf{0.850}}             & \textcolor{myRed}{\textbf{0.888}}         & \textcolor{myRed}{\textbf{0.787}}         & \textcolor{myRed}{\textbf{0.045}}    \\
\bottomrule
\end{tabular}
\label{supp_noisy}
\end{table}
\begin{table*}[t]
\centering
\caption{Performance of the proposed model under various backbones. }
\label{fig:backbone}
\fontsize{6.5}{9}\selectfont
\begin{tabular}{cc|p{1.2cm}|*{3}{p{0.6cm}}<{\centering}p{0.8cm} |*{3}{p{0.6cm}}<{\centering}p{0.8cm}|*{3}{p{0.6cm}}<{\centering}p{0.8cm}}
\toprule
\multicolumn{2}{c|}{Condition}&\multicolumn{1}{c|}{\multirow{2}{*}{Backbone}}& \multicolumn{4}{c|}{VT821} & \multicolumn{4}{c|}{VT1000} & \multicolumn{4}{c}{VT5000} \\
\multicolumn{1}{c}{RGB}&\multicolumn{1}{c|}{T}&& $S_\alpha$$\uparrow$ & $E_m$$\uparrow$ & $F_\beta$$\uparrow$ & MAE$\downarrow$ & $S_\alpha$$\uparrow$ & $E_m$$\uparrow$ & $F_\beta$$\uparrow$ & MAE$\downarrow$ & $S_\alpha$$\uparrow$ & $E_m$$\uparrow$ & $F_\beta$$\uparrow$ & MAE$\downarrow$ \\
\midrule
% \multicolumn{17}{l}{\textbf{Performance of RGB-D Models under Modality-Complete}}\\
\multirow{5}[1]{*}{\solidcircle} &\multirow{5}[1]{*}{\solidcircle}
&ResNet50 & 0.897 & 0.923 & 0.852 & 0.031 & 0.928 & 0.956 & 0.896 & 0.024 & 0.892 & 0.927 & 0.843 & 0.037 \\
&&ResNet101 & 0.904 & 0.927 & 0.858 & 0.030 & 0.929 & 0.953 & 0.890 & 0.024 & 0.893 & 0.929 & 0.844 & 0.036 \\
&&Swin-T & 0.899 & 0.920 & 0.850 & 0.030 & 0.927 & 0.955 & 0.895 & 0.026 & 0.894 & 0.930 & 0.853 & 0.036 \\
&&Swin-S & 0.899 & 0.922 & 0.854 & 0.028 & 0.929 & 0.958 & 0.893 & 0.023 & 0.895 & 0.934 & 0.850 & 0.034 \\
&&Swin-B & \textcolor{myRed}{0.905} & \textcolor{myRed}{0.930} & \textcolor{myRed}{0.860} & \textcolor{myRed}{0.026} & \textcolor{myRed}{0.935} & \textcolor{myRed}{0.962} & \textcolor{myRed}{0.901} & \textcolor{myRed}{0.020} & \textcolor{myRed}{0.906} & \textcolor{myRed}{0.941} & \textcolor{myRed}{0.857} & \textcolor{myRed}{0.028}
 \\

\midrule
% \multicolumn{17}{l}{\textbf{Performance of RGB-D Models under Missing RGB}}\\
\multirow{5}[1]{*}{\emptycircle} &\multirow{5}[1]{*}{\solidcircle}
& Resnet50 & 0.809 & 0.865 & 0.757 & 0.053 & 0.886 & 0.930 & 0.858 & 0.041 & 0.831 & 0.885 & 0.774 & 0.052 \\
&&Resnet101 & 0.811 & 0.860 & 0.749 & 0.051 & 0.889 & 0.929 & 0.861 & 0.042 & 0.829 & 0.882 & 0.765 & 0.053 \\
&&Swin-T & 0.808 & 0.862 & 0.750 & 0.055 & 0.885 & 0.919 & 0.847 & 0.047 & 0.825 & 0.872 & 0.770 & 0.055 \\
&&Swin-S & 0.820 & 0.867 & 0.755 & 0.056 & 0.890 & 0.928 & 0.855 & 0.039 & 0.840 & 0.885 & 0.769 & 0.054 \\
&&Swin-B & \textcolor{myRed}{0.824} & \textcolor{myRed}{0.874} & \textcolor{myRed}{0.766} & \textcolor{myRed}{0.049} & \textcolor{myRed}{0.895} & \textcolor{myRed}{0.935} & \textcolor{myRed}{0.863} & \textcolor{myRed}{0.040} & \textcolor{myRed}{0.840} & \textcolor{myRed}{0.894} & \textcolor{myRed}{0.782} & \textcolor{myRed}{0.047}
\\
\midrule
\multirow{5}[1]{*}{\solidcircle} &\multirow{5}[1]{*}{\emptycircle}
% & SPNet & 0.857 & 0.910 & 0.864 & 0.063 & 0.913 & 0.953 & 0.900 & 0.027 & 0.920 & 0.968 & 0.909 & 0.021 & 0.898 & 0.938 & 0.903 & 0.045 & 0.903 & 0.943 & 0.899 & 0.040\\
&Resnet50 & 0.872 & 0.900 & 0.816 & 0.043 & 0.914 & 0.943 & 0.882 & 0.030 & 0.874 & 0.914 & 0.822 & 0.043 \\
% & &SPNet & 0.570 & 0.707 & 0.537 & 0.178 & 0.631 & 0.766 & 0.517 & 0.106 & 0.704 & 0.753 & 0.636 & 0.089 & 0.587 & 0.723 & 0.560 & 0.175 & 0.454 & 0.602 & 0.366 & 0.200\\
&&Resnet101 & 0.877 & 0.895 & 0.811 & 0.042 & 0.916 & 0.940 & 0.884 & 0.031 & 0.872 & 0.910 & 0.817 & 0.045 \\
&&Swin-T & 0.869 & 0.889 & 0.805 & 0.045 & 0.908 & 0.933 & 0.873 & 0.035 & 0.865 & 0.905 & 0.820 & 0.047 \\
&&Swin-S & 0.880 & 0.900 & 0.815 & 0.039 & 0.920 & 0.939 & 0.880 & 0.028 & 0.880 & 0.915 & 0.830 & 0.042 \\
&&Swin-B & \textcolor{myRed}{0.883} & \textcolor{myRed}{0.906} & \textcolor{myRed}{0.828} & \textcolor{myRed}{0.037} & \textcolor{myRed}{0.926} & \textcolor{myRed}{0.952} & \textcolor{myRed}{0.890} & \textcolor{myRed}{0.025} & \textcolor{myRed}{0.885} & \textcolor{myRed}{0.922} & \textcolor{myRed}{0.832} & \textcolor{myRed}{0.038}
 \\
\bottomrule
\end{tabular}
\label{supp_rgb-d}
\end{table*}

\subsection{Evaluation Metrics of Average and Average dropout}
\label{sec:Average and Average dropout}
In this section, we introduce two metrics mentioned in the main text, \textbf{Average} and \textbf{Average Dropout}. Their calculations are simple, but they intuitively reflect the overall performance of the model in situations of modal absence and completeness, demonstrating the strength or weakness of the model's robustness.

\textbf{Average.} For each metric $F$, we calculate its scores $F_{full}$, $F_{m1}$, $F_{m2}$ under modality-complete, missing $m_2$ and missing $m_1$. Then we simply compute their average:
\begin{equation}
    \mathcal{A} = \frac{F_{full} + F_{m1} + F_{m2}}{3},
\end{equation}
$\mathcal{A}$ is the calculated metric \textbf{Average}, which represents the comprehensive performance of the model under both missing and complete conditions.

\textbf{Average dropout.} For Average Dropout, we aim to reflect the performance degradation of the model under missing conditions relative to the full modality performance. Thus, we adopt the following calculation method:
\begin{equation}
    \mathcal{D} = \frac{(F_{m1} - F_{full}) + (F_{m2} - F_{full})}{2},
\end{equation}
where $\mathcal{D}$ denotes $\textbf{Average Dropout}$. As can be seen, if the model suffers significant performance loss under the missing modality, this can be reflected through $\textbf{Average Dropout}$.

\subsection{Experiments on Various Backbones}
\label{sec:Experiments on Various Backbones}
In our study, we have conducted a series of experiments to evaluate the performance of prevalent backbone architectures in the context of saliency detection, which includes ResNet50 \cite{he2016deep}, ResNet101 \cite{he2016deep}, and Swin Transformer \cite{liu2021swin} variants (Swin-T, Swin-S, Swin-B). We performed these evaluations under conditions of both complete and missing modalities as shown in Table \ref{fig:backbone}, encompassing all three datasets within the RGB-T scope. It can be observed that compared to ResNet50, ResNet101 has shown some improvements in results. Within the experiments involving Swin architectures, Swin-B achieved the best outcomes, although Swin-T and Swin-S did not exhibit significant changes compared to ResNet50. The systematic experimentation across different backbones and modalities aims to discern the robustness and adaptability of our framework.
\subsection{Experiments on Noisy Inputs}
\label{sec:Experiments on Noisy Inputs}
% The existing RGB-T and RGB-D datasets share a common issue: the introduction of dual-modal aims to address the recognition difficulties caused by insufficient information in single modality. However, in current datasets, only a few samples consider the utility of thermal or depth image. 
\textbf{VT821-Noisy.} To better assess the effectiveness of our proposed LQA module, we aggregate pairs of images with inherent noise from all RGB images in the VT821 dataset (referred to as VT821-Noisy), \textbf{which is a sub-set of the original VT821}.

\begin{table*}[t]
\caption{Quantitative experiments of different RGB-T models trained by Modality Dropout in modality-complete and modality-missing conditions. $\uparrow$ denotes that a larger value is better. The best results are in red, the second-best results are in blue, and the third-best results are in green.}
% \label{pe-c-surf}
\centering
\label{supp_comparison with others in MD}
\fontsize{6.5}{9}\selectfont
\begin{tabular}{c|p{0.8cm}<{\centering}p{0.6cm}<{\centering}|c|*{6}{p{0.9cm}<{\centering}}p{1.1cm}<{\centering}|c}
\toprule
    \multirow{2}{*}{Datasets} & \multicolumn{2}{c|}{Conditions} & \multirow{2}{*}{Metric}  & CSRNet   & ADF & TNet &DCNet & LSNet &MIDD     &TAGFNet  & Ours\\
\cmidrule{2-3}         & \multicolumn{1}{c|}{RGB} & T     &          & \cite{huo2021efficient}  & \cite{tu2022rgbt} &\cite{cong2022does} &\cite{tu2022weakly} &\cite{zhou2023lsnet} &\cite{tu2021multi}    &\cite{wang2023thermal} &\\
    \midrule
     \multirow{10}{*}{VT821} & \multirow{2}[1]{*}{\solidcircle} & \multirow{2}[1]{*}{\solidcircle} & 
    \textit{$E_m$$\uparrow$}  &0.840 & 0.847 & 0.900 & \textcolor{myBlue}{0.911} & \textcolor{myBlue}{0.911} & \textcolor{myGreen}{0.903} & 0.886 
    &\textcolor{myRed}{\textbf{0.922}}\\
  &       &       & 
    \textit{$F_\beta$$\uparrow$}  &0.717 & 0.752 & 0.820 & \textcolor{myBlue}{0.834} & 0.815 & \textcolor{myGreen}{0.822} & 0.803 & \textcolor{myRed}{\textbf{0.849}}\\
  & \multirow{2}[0]{*}{\emptycircle} & \multirow{2}[0]{*}{\solidcircle} & 
    \textit{$E_m$$\uparrow$}  &0.727 & 0.785 & 0.815 & \textcolor{myGreen}{0.846} & \textcolor{myBlue}{0.859} & 0.804 & 0.782   &\textcolor{myRed}{\textbf{0.864}}\\
  &       &       & 
    \textit{$F_\beta$$\uparrow$}  & 
    0.558 & 0.653 & 0.720 & \textcolor{myBlue}{0.731} & \textcolor{myGreen}{0.725} & 0.683 & 0.636 &\textcolor{myRed}{\textbf{0.752}} \\
  & \multirow{2}[1]{*}{\solidcircle} & \multirow{2}[1]{*}{\emptycircle} 
   & 
    \textit{$E_m$$\uparrow$}  & 0.763 & 0.809 & 0.847 & \textcolor{myBlue}{0.885} & \textcolor{myGreen}{0.863} & 0.855 & 0.802 & \textcolor{myRed}{\textbf{0.900}}\\
  &       &       & 
    \textit{$F_\beta$$\uparrow$}  & 0.595 & 0.701 & \textcolor{myGreen}{0.761} & \textcolor{myBlue}{0.788} & 0.760 & 0.758 & 0.671  & \textcolor{myRed}{\textbf{0.817}}\\
\cmidrule{2-12}          & \multicolumn{2}{c|}{\multirow{2}{*}{Average Drop}} & 
    \textit{\(E_m\)\(\uparrow\)}  & -0.095 & \textcolor{myGreen}{-0.050} & -0.059 & \textcolor{myBlue}{-0.046} & \textcolor{myRed}{\textbf{-0.040}} & -0.074 & -0.094  & \textcolor{myRed}{\textbf{-0.040}} \\
  & \multicolumn{2}{c|}{} &\textit{$F_\beta$$\uparrow$}& 
    -0.140 & -0.075 & \textcolor{myBlue}{-0.069} & -0.074 & \textcolor{myGreen}{-0.073} & -0.102 & -0.150 & \textcolor{myRed}{\textbf{-0.065}} \\      
    
    & \multicolumn{2}{c|}{\multirow{2}{*}{Average}} & 
    \textit{$E_m$$\uparrow$}   & 0.776  & 0.814  & 0.854  & \textcolor{myBlue}{0.881}  & \textcolor{myGreen}{0.874}  & 0.854  & 0.823   & \textcolor{myRed}{\textbf{0.895}} \\
  & \multicolumn{2}{c|}{} & 
    \textit{$F_\beta$$\uparrow$}  & 0.623  & 0.702  & \textcolor{myGreen}{0.775}  & \textcolor{myBlue}{0.784}  & 0.770  & 0.754  & 0.703    & \textcolor{myRed}{\textbf{0.806}} \\
    \midrule

    % VT1000数据集数据
    \multirow{10}{*}{VT1000} & \multirow{2}[1]{*}{\solidcircle} & \multirow{2}[1]{*}{\solidcircle} & 
      \textit{$E_m$$\uparrow$}  & 0.902 & 0.923 & 0.925 & 0.937 & \textcolor{myBlue}{0.939} & \textcolor{myGreen}{0.937} & 0.929 & \textcolor{myRed}{\textbf{0.955}}\\
      &       &       & 
      \textit{$F_\beta$$\uparrow$}  & 0.815 & 0.853 & 0.861 & \textcolor{myBlue}{0.881} & 0.870 & \textcolor{myGreen}{0.876} & 0.859  & \textcolor{myRed}{\textbf{0.895}}\\
      & \multirow{2}[0]{*}{\emptycircle} & \multirow{2}[0]{*}{\solidcircle} & 
      \textit{$E_m$$\uparrow$}  & 0.806 & 0.865 & 0.884 & \textcolor{myBlue}{0.918} & \textcolor{myGreen}{0.916} & 0.892 & 0.818 & \textcolor{myRed}{\textbf{0.930}}\\
      &       &       & 
      \textit{$F_\beta$$\uparrow$}  &  0.667 & 0.773 & 0.810 & \textcolor{myBlue}{0.837} & 0.809 & \textcolor{myGreen}{0.812} & 0.674 & \textcolor{myRed}{\textbf{0.855}}\\
      & \multirow{2}[1]{*}{\solidcircle} & \multirow{2}[1]{*}{\emptycircle} 
      & 
    \textit{$E_m$$\uparrow$}  &0.810 & 0.902 & \textcolor{myGreen}{0.910} & 0.907 & \textcolor{myBlue}{0.915} & 0.904 & 0.824  & \textcolor{myRed}{\textbf{0.943}}\\
          &       &       & 
      \textit{$F_\beta$$\uparrow$}  & 0.691 & 0.830 & 0.845 & \textcolor{myBlue}{0.867} & 0.850 & \textcolor{myGreen}{0.862} & 0.692& \textcolor{myRed}{\textbf{0.881}} \\
\cmidrule{2-12}          & \multicolumn{2}{c|}{\multirow{2}{*}{Average Drop}} & 
    \textit{\(E_m\)\(\uparrow\)}   & -0.094 & -0.040 & -0.028 & \textcolor{myGreen}{-0.025} & \textcolor{myBlue}{-0.024} & -0.039 & -0.108 & \textcolor{myRed}{\textbf{-0.018}}\\
  & \multicolumn{2}{c|}{} & 
    \textit{\(F_\beta\)\(\uparrow\)}  & --0.136 & -0.052 & \textcolor{myGreen}{-0.034} & \textcolor{myBlue}{-0.029} & -0.041 & -0.039 & -0.176& \textcolor{myRed}{\textbf{-0.027}}\\
      & \multicolumn{2}{c|}{\multirow{2}{*}{Average}} & 
      \textit{$E_m$$\uparrow$}   & 0.839 & 0.897 & 0.906 & \textcolor{myGreen}{0.921} & \textcolor{myBlue}{0.923} & 0.911 & 0.857   & \textcolor{myRed}{\textbf{0.943}} \\
      & \multicolumn{2}{c|}{} & 
      \textit{$F_\beta$$\uparrow$}  & 0.724 & 0.818 & 0.836 & \textcolor{myBlue}{0.862} & 0.843 & \textcolor{myGreen}{0.850} & 0.742  & \textcolor{myRed}{\textbf{0.877}} \\
\midrule
% VT5000数据集数据.
\multirow{10}{*}{VT5000} & \multirow{2}[1]{*}{\solidcircle} & \multirow{2}[1]{*}{\solidcircle} & 
    \textit{$E_m$$\uparrow$}  & 0.828 & 0.874 & \textcolor{myBlue}{0.910} & \textcolor{myGreen}{0.903} & 0.902 & 0.897 & 0.891 & \textcolor{myRed}{\textbf{0.927}} \\
          &       &       & 
    \textit{$F_\beta$$\uparrow$}  & 0.677 & 0.764 & \textcolor{myBlue}{0.815} & \textcolor{myGreen}{0.812} & 0.795 & 0.798 & 0.796  & \textcolor{myRed}{\textbf{0.843}} \\
    & \multirow{2}[0]{*}{\emptycircle} & \multirow{2}[0]{*}{\solidcircle} & 
    \textit{$E_m$$\uparrow$}  & 0.743 & 0.802 & 0.822 & \textcolor{myGreen}{0.854} & \textcolor{myBlue}{0.857} & 0.823 & 0.784  & \textcolor{myRed}{\textbf{0.887}} \\
          &       &       & 
    \textit{$F_\beta$$\uparrow$}  & 0.536 & 0.660 & \textcolor{myGreen}{0.746} & \textcolor{myBlue}{0.748} & 0.728 & 0.701 & 0.614  & \textcolor{myRed}{\textbf{0.774}} \\
    
    & \multirow{2}[1]{*}{\solidcircle} & \multirow{2}[1]{*}{\emptycircle} 
    & 
    \textit{$E_m$$\uparrow$}   & 0.750 & 0.853 & 0.865 & \textcolor{myGreen}{0.880} & \textcolor{myBlue}{0.885} & \textcolor{myGreen}{0.880} & 0.802  & \textcolor{myRed}{\textbf{0.913}} \\ &       &       & 
    \textit{$F_\beta$$\uparrow$}  & 0.555 & 0.736 & 0.756 & \textcolor{myBlue}{0.781} & 0.768 & \textcolor{myGreen}{0.773} & 0.652  & \textcolor{myRed}{\textbf{0.822}} \\
\cmidrule{2-12}          & \multicolumn{2}{c|}{\multirow{2}{*}{Average Drop}} & 
    \textit{\(E_m\)\(\uparrow\)}   & -0.082 & -0.047 & -0.067 & \textcolor{myGreen}{-0.036} & \textcolor{myBlue}{-0.031} & -0.046 & -0.098 & \textcolor{myRed}{\textbf{-0.027}} \\
  & \multicolumn{2}{c|}{} & 
    \textit{\(F_\beta\)\(\uparrow\)}  & -0.131 & -0.066 & -0.064& \textcolor{myGreen}{-0.048} & \textcolor{myBlue}{-0.047} & -0.061 & -0.163& \textcolor{myRed}{\textbf{-0.045}} \\
          & \multicolumn{2}{c|}{\multirow{2}{*}{Average}} &
    \textit{$E_m$$\uparrow$}   & 0.773  & 0.843  & 0.866  & \textcolor{myGreen}{0.879}  &\textcolor{myBlue}{0.881}  & 0.867  & 0.826    & \textcolor{myRed}{\textbf{0.909}}  \\
          & \multicolumn{2}{c|}{} & 
          \textit{$F_\beta$$\uparrow$}   & 0.589  & 0.720  & \textcolor{myGreen}{0.772}  & \textcolor{myBlue}{0.780}  & 0.764  & 0.757  & 0.687    & \textcolor{myRed}{\textbf{0.813}}  \\
    \bottomrule
\end{tabular}

\end{table*}
% \begin{figure}[t]
% \centering
% \includegraphics[width=1\linewidth,keepaspectratio]{img/BRISQUE1.4.pdf}
% \caption{The score calculated using BRISQUE and LQA.}
% \label{fig:brisque}
% \end{figure}
%这里做点可视化

\begin{figure*}[ht!]
\centering
\includegraphics[width=1\linewidth,keepaspectratio]{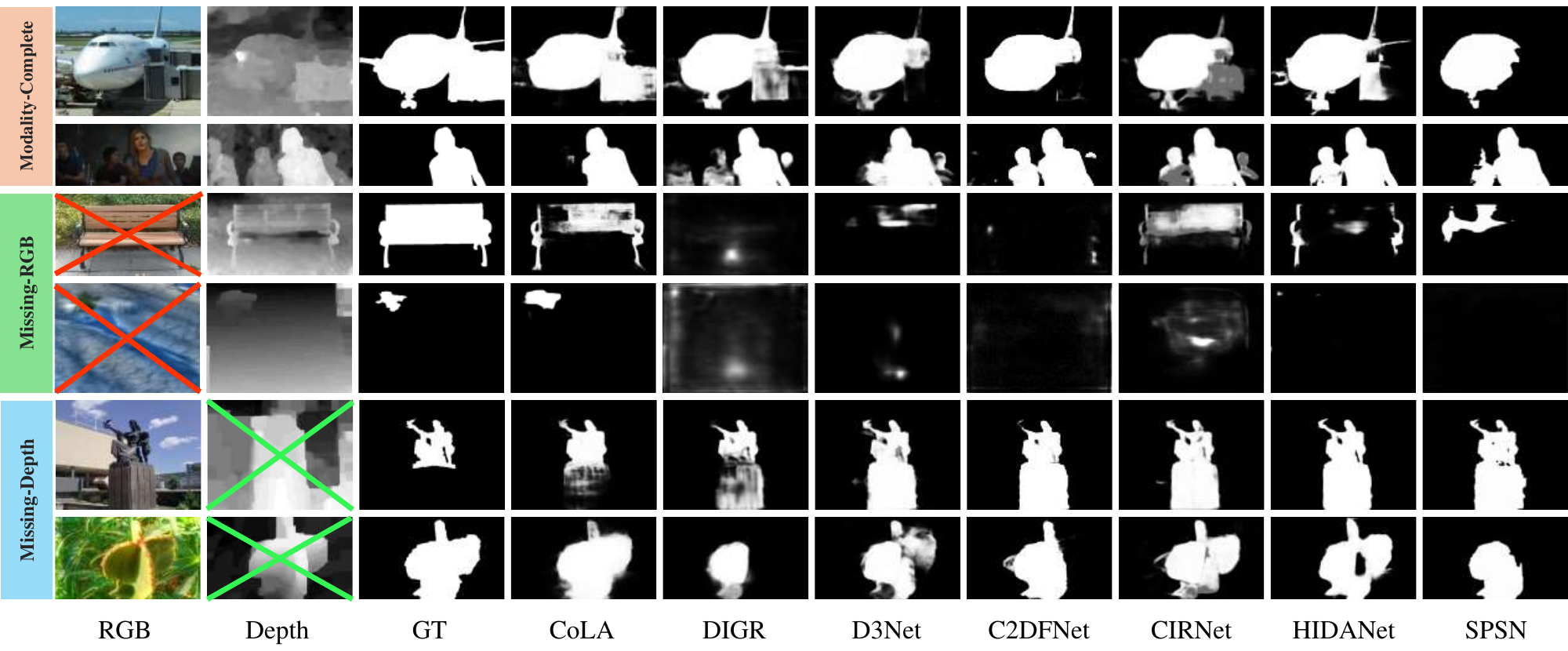}
\caption{Qualitative comparisons of the proposed model and existing state-of-the-art methods in three situations. The red or green crosses on the image indicate that RGB or Depth modality is missing. }
\label{supp_visual_depth}
\end{figure*}
\textbf{Evaluation.} We tested various RGB-T methods and our method for their performance on the full modality of VT821-noisy. As shown in Table \ref{supp_noisy}, compared to the test results with the complete VT821 dataset (Table \textcolor{red}{1}), our model demonstrated greater advantages, indicating that our model has more significant effects when dealing with noisy images.
\section{More Experiments of RGB-D inputs}
\label{sec:More Experiments of RGB-D inputs}
\subsection{Quantitative Evaluation}
\label{sec:Quantitative Evaluation}

Our RGB-D quantitative comparison results are shown in Table \ref{supp_rgb-d} and \ref{supp_rgb-d-2}. The compared methods include D3Net \cite{fan2020rethinking}, DIGR \cite{cheng2022depth}, C$^{2}$DFNet \cite{zhang2022c}, CIRNet \cite{cong2022cir}, SPSN \cite{lee2022spsn}, HiDAnet\cite{wu2023hidanet}. Compared to RGB-T, the performance degradation of RGB-D methods in the missing of depth is significantly smaller than when RGB is missing. This indicates that in the RGB-D domain, RGB holds a dominant position over depth, and therefore, comparative methods experience significant performance loss when the RGB modality is missing. Our model achieves near-optimal performance in both missing modalities. With evaluation metrics that reflect the model's robustness and overall capability reaching the highest levels in Average and Average Drop metrics. This demonstrates that our model has the best robustness and comprehensive capability.

\begin{table*}[t]
\centering
\caption{Validation experiment results of different RGB-D models in modality-complete and modality-missing conditions with SIP,NLPR and DES datasets. }
\fontsize{5}{9}\selectfont
\begin{tabular}{cc|c|cccc|cccc|cccc}
\toprule
\multicolumn{2}{c|}{Condition}&\multicolumn{1}{c|}{\multirow{2}{*}{Method}}& \multicolumn{4}{c|}{SIP} & \multicolumn{4}{c|}{NLPR} & \multicolumn{4}{c}{DES} \\
\multicolumn{1}{c}{RGB}&\multicolumn{1}{c|}{Depth}&& $S_\alpha$$\uparrow$ & $E_m$$\uparrow$ & $F_\beta$$\uparrow$ & MAE$\downarrow$ & $S_\alpha$$\uparrow$ & $E_m$$\uparrow$ & $F_\beta$$\uparrow$ & MAE$\downarrow$ & $S_\alpha$$\uparrow$ & $E_m$$\uparrow$ & $F_\beta$$\uparrow$ &MAE$\downarrow$\\
\midrule
% \multicolumn{17}{l}{\textbf{Performance of RGB-D Models under Modality-Complete}}\\
\multirow{6}[1]{*}{\solidcircle} &\multirow{6}[1]{*}{\solidcircle}
&D3Net&0.860&0.895&0.844&0.061&0.912&0.944&0.887&0.030&0.898&0.938&0.880&0.031\\
% && SPNet & 0.894 & 0.925 & 0.885 & 0.043& 0.921 & 0.950 & 0.889 & 0.024& 0.945 & 0.980 & 0. & 0.028& 0.887 & 0.930 & 0.887 & 0.047& 0.897 & 0.942 & 0.895 & 0.041\\
&&DIGR & 0.882 & 0.927 & 0.881 & 0.051 & 0.931 & 0.950 & 0.896 & \textcolor{myRed}{\textbf{0.020}} & 0.936 & 0.962 & 0.922 & 0.019 \\
&&C2DFNet& 0.872 & 0.910 & 0.866 & 0.054 & 0.927 & \textcolor{myRed}{\textbf{0.962}} & 0.907 & 0.022 & 0.919 & 0.940 & 0.905 & 0.020 \\
&&CIRNet:& 0.888 & 0.907 & 0.855 & 0.053 & \textcolor{myBlue}{0.933} & 0.949 & 0.897 & 0.023 & 0.932 & 0.943 & 0.905 & 0.025 \\
&&SPSN & 0.890 & 0.914 & \textcolor{myBlue}{0.896} & 0.042 & 0.926 & 0.946 & \textcolor{myRed}{\textbf{0.914}} & 0.022 & \textcolor{myBlue}{0.938} & 0.950 & \textcolor{myRed}{\textbf{0.943}} & \textcolor{myBlue}{0.016} \\
&&HIDANet: & \textcolor{myRed}{\textbf{0.899}} & \textcolor{myBlue}{0.932} & 0.893 & \textcolor{myRed}{\textbf{0.040}} & 0.930 & \textcolor{myBlue}{0.959} & \textcolor{myBlue}{0.909} & 0.022 & \textcolor{myRed}{\textbf{0.947}} & \textcolor{myRed}{\textbf{0.979}} & \textcolor{myBlue}{0.941} & \textcolor{myRed}{\textbf{0.014}} 
\\
\cmidrule{3-15} 
&&Ours: &  \textcolor{myBlue}{0.895} & \textcolor{myRed}{\textbf{0.935}} & \textcolor{myRed}{\textbf{0.894}} & \textcolor{myBlue}{0.042} & \textcolor{myRed}{\textbf{0.935}} & 0.957 & \textcolor{myBlue}{0.909} & \textcolor{myBlue}{0.021} & 0.935 & \textcolor{myBlue}{0.963} & 0.925 & 0.018 \\
\midrule
% \multicolumn{17}{l}{\textbf{Performance of RGB-D Models under Missing RGB}}\\
\multirow{6}[1]{*}{\emptycircle} &\multirow{6}[1]{*}{\solidcircle}
&D3Net&0.720&0.699&0.689&0.112&0.742&0.730&0.695&0.099&0.758&0.774&0.725&0.105\\
&&DIGR&0.699&0.635&0.534&0.146&0.661&0.603&0.422&0.119&0.668&0.586&0.411&0.122\\
&&C2DFNet&0.687&0.606&0.538&0.124&0.621&0.560&0.538&0.114&0.604&0.568&0.509&0.106\\
&&CIRNet&0.787&0.750&0.721&0.115&0.847&0.823&0.753&0.061&\textcolor{myBlue}{0.893}&0.867&0.814&0.047\\
&&SPSN&0.740&0.758&0.725&0.106&0.709&0.700&0.599&0.070&0.768&0.793&0.731&0.051\\
&&HIDANet&\textcolor{myBlue}{0.813}&\textcolor{myBlue}{0.854}&\textcolor{myBlue}{0.805}&\textcolor{myBlue}{0.077}&\textcolor{myBlue}{0.823}&\textcolor{myBlue}{0.878}&\textcolor{myBlue}{0.779}&\textcolor{myBlue}{0.063}&0.857&\textcolor{myBlue}{0.907}&\textcolor{myBlue}{0.832}&\textcolor{myBlue}{0.050}
\\
\cmidrule{3-15} 
&&Ours&\textcolor{myRed}{\textbf{0.843}} & \textcolor{myRed}{\textbf{0.863}} & \textcolor{myRed}{\textbf{0.822}} & \textcolor{myRed}{\textbf{0.076}} & \textcolor{myRed}{\textbf{0.865}} & \textcolor{myRed}{\textbf{0.903}} & \textcolor{myRed}{\textbf{0.809}} & \textcolor{myRed}{\textbf{0.053}} & \textcolor{myRed}{\textbf{0.909}} & \textcolor{myRed}{\textbf{0.947}} & \textcolor{myRed}{\textbf{0.874}} & \textcolor{myRed}{\textbf{0.030}}\\
\midrule
\multirow{6}[1]{*}{\solidcircle} &\multirow{6}[1]{*}{\emptycircle}
% & SPNet & 0.857 & 0.910 & 0.864 & 0.063 & 0.913 & 0.953 & 0.900 & 0.027 & 0.920 & 0.968 & 0.909 & 0.021 & 0.898 & 0.938 & 0.903 & 0.045 & 0.903 & 0.943 & 0.899 & 0.040\\
& D3Net & 0.831 & 0.871 & 0.795 & 0.079 & 0.903 & 0.923 & 0.855 & 0.033 & 0.870 & 0.874 & 0.811 & 0.040 \\
& &DIGR & 0.797 & 0.824 & 0.779 & 0.086 & 0.887 & 0.902 & 0.847 & 0.036 & 0.850 & 0.837 & 0.790 & 0.042\\
& &C2DFNet & 0.812 & 0.858 & 0.800 & 0.077 & \textcolor{myBlue}{0.911} & \textcolor{myBlue}{0.940} & \textcolor{myBlue}{0.882}& \textcolor{myBlue}{0.031} & 0.882 & 0.902& 0.856 & 0.029\\
& &CIRNet & 0.847 & 0.869 & 0.825 & 0.076 &\textcolor{myBlue}{0.911} & 0.925 & 0.876 & 0.035 & 0.873 & 0.860 & 0.825 & 0.042\\
&& SPSN & 0.844 & 0.894 & 0.837 & 0.065 & 0.902 & 0.934 & 0.868 & \textcolor{myBlue}{0.031} & 0.884 & 0.908 & 0.854 & 0.033\\
& &HIDANet & \textcolor{myBlue}{0.856} & \textcolor{myBlue}{0.900} & \textcolor{myBlue}{0.848} & \textcolor{myBlue}{0.063} & 0.907 & 0.938 & 0.877 & 0.032 & \textcolor{myBlue}{0.895} & \textcolor{myBlue}{0.922} & \textcolor{myBlue}{0.875} & \textcolor{myBlue}{0.031}\\
\cmidrule{3-15} 
&& Ours & \textcolor{myRed}{\textbf{0.863}} & \textcolor{myRed}{\textbf{0.908}} & \textcolor{myRed}{\textbf{0.855}} & \textcolor{myRed}{\textbf{0.057}} & \textcolor{myRed}{\textbf{0.916}} & \textcolor{myRed}{\textbf{0.946}} & \textcolor{myRed}{\textbf{0.885}} & \textcolor{myRed}{\textbf{0.029}} & \textcolor{myRed}{\textbf{0.900}} & \textcolor{myRed}{\textbf{0.926}}& \textcolor{myRed}{\textbf{0.880}}& \textcolor{myRed}{\textbf{0.030}}\\
\midrule
\multicolumn{2}{c|}{\multirow{7}{*}{\textbf{Average}}}
&D3Net & 0.804 & 0.822 & 0.776 & 0.084 & 0.852 & 0.866 & 0.812 & 0.054 & 0.842 & 0.862 & 0.805 & 0.059 \\
&&DIGR & 0.793 & 0.795 & 0.731 & 0.094 & 0.826 & 0.818 & 0.722 & 0.059 & 0.818 & 0.795 & 0.708 & 0.061\\
&&C2DFNet & 0.790 & 0.791 & 0.735 & 0.085 & 0.820 & 0.821 & 0.776 & 0.056 & 0.802 & 0.803 & 0.757 & 0.052\\
&&CIRNet & 0.841 & 0.842 & 0.800 & 0.081 & 0.897 & 0.899 & 0.842 & 0.040 & 0.899 & 0.890 & 0.848 & 0.038\\
&&SPSN & 0.825 & 0.855 & 0.819 & 0.071 & 0.846 & 0.860 & 0.794 & 0.041 & 0.863 & 0.884 & 0.843 & 0.033\\
&&HIDANet & \textcolor{myBlue}{0.862} & \textcolor{myBlue}{0.895} & \textcolor{myBlue}{0.849} & \textcolor{myBlue}{0.060} & \textcolor{myBlue}{0.887} & \textcolor{myBlue}{0.925} & \textcolor{myBlue}{0.855} & \textcolor{myBlue}{0.039} & \textcolor{myBlue}{0.900} & \textcolor{myBlue}{0.936} & \textcolor{myBlue}{0.883} & \textcolor{myBlue}{0.032}\\
\cmidrule{3-15} 
&&Ours  & \textcolor{myRed}{\textbf{0.867}}  & \textcolor{myRed}{\textbf{0.902}}  & \textcolor{myRed}{\textbf{0.857}}  & \textcolor{myRed}{\textbf{0.058}}  & \textcolor{myRed}{\textbf{0.905}}  & \textcolor{myRed}{\textbf{0.935}}  & \textcolor{myRed}{\textbf{0.868}}  & \textcolor{myRed}{\textbf{0.034}}  & \textcolor{myRed}{\textbf{0.915}}  & \textcolor{myRed}{\textbf{0.945}}  & \textcolor{myRed}{\textbf{0.893}}  & \textcolor{myRed}{\textbf{0.026}}  
\\
\midrule
\multicolumn{2}{c|}{\multirow{7}{*}{ \textbf{Average Drop}}}
&D3Net & -0.085 & -0.110 & -0.102 & 0.035 & -0.090 & -0.118 & -0.112 & 0.036 & -0.084 & -0.114 & -0.112 & 0.042\\
&&DIGR & -0.134 & -0.198 & -0.225 & 0.065 & -0.157 & -0.198 & -0.262 & 0.057 & -0.177 & -0.251 & -0.322 & 0.063\\
&&C2DFNet & -0.123 & -0.178 & -0.197 & 0.047 & -0.161 & -0.212 & -0.197 & 0.051 & -0.176 & -0.205 & -0.223 & 0.048\\
&&CIRNet & -0.071 & -0.098 & -0.082 & 0.043 & \textcolor{myBlue}{-0.054} & -0.075 & -0.083 & \textcolor{myBlue}{0.025} & \textcolor{myBlue}{-0.049} & -0.080 & \textcolor{myBlue}{-0.086} & \textcolor{myBlue}{0.020}\\
&&SPSN & -0.098 & -0.088 & -0.115 & 0.044 & -0.121 & -0.129 & -0.181 & 0.029 & -0.112 & -0.099 & -0.151 & 0.026\\
&&HIDANet & \textcolor{myBlue}{-0.055} & \textcolor{myBlue}{-0.055} & \textcolor{myBlue}{-0.067} & \textcolor{myBlue}{0.030} &-0.065 & \textcolor{myBlue}{-0.051} & \textcolor{myBlue}{-0.081} & 0.026 &-0.071 & \textcolor{myBlue}{-0.065} & -0.088 &0.027
\\
    \cmidrule{3-15} 
    &&Ours  & \textcolor{myRed}{\textbf{-0.042}}  & \textcolor{myRed}{\textbf{-0.050}}  & \textcolor{myRed}{\textbf{-0.056}}  & \textcolor{myRed}{\textbf{0.025}}  & \textcolor{myRed}{\textbf{-0.045}}  & \textcolor{myRed}{\textbf{-0.033}}  & \textcolor{myRed}{\textbf{-0.062}}  & \textcolor{myRed}{\textbf{0.020}}  & \textcolor{myRed}{\textbf{-0.031}}  & \textcolor{myRed}{\textbf{-0.027}}  & \textcolor{myRed}{\textbf{-0.048}}  & \textcolor{myRed}{\textbf{0.012}} 
 \\

\bottomrule
\end{tabular}
\label{supp_rgb-d}
\end{table*}
\begin{table*}[t]
\centering
\caption{Validation experiment results of different RGB-D models in modality-complete and modality-missing conditions with NJU2K and STERE datasets. }
\fontsize{5}{9}\selectfont
\begin{tabular}{cc|c|cccc|cccc}
\toprule
\multicolumn{2}{c|}{Condition}&\multicolumn{1}{c|}{\multirow{2}{*}{Method}}& \multicolumn{4}{c|}{NJU2K} & \multicolumn{4}{c}{STERE}\\
\multicolumn{1}{c}{RGB}&\multicolumn{1}{c|}{Depth}&& $S_\alpha$$\uparrow$ & $E_m$$\uparrow$ & $F_\beta$$\uparrow$ & MAE$\downarrow$ & $S_\alpha$$\uparrow$ & $E_m$$\uparrow$ & $F_\beta$$\uparrow$ & MAE$\downarrow$\\
\midrule
% \multicolumn{17}{l}{\textbf{Performance of RGB-D Models under Modality-Complete}}\\
\multirow{6}[1]{*}{\solidcircle} &\multirow{6}[1]{*}{\solidcircle}
&D3Net&0.900&0.921&0.890&0.041&0.899&0.925&0.879&0.041\\
&&DIGR&\textcolor{myBlue}{0.933}&0.941&0.900&\textcolor{myRed}{\textbf{0.026}}&\textcolor{myRed}{\textbf{0.914}}&0.934&0.880&\textcolor{myBlue}{0.035}\\
&&C2DFNet&0.907&0.936&0.898&0.039&0.903&0.937&0.884&0.038\\
&&CIRNet&0.925&0.933&0.895&0.035&\textcolor{myRed}{\textbf{0.914}}&0.930&0.874&0.038\\
&&SPSN&0.912&0.936&0.912&0.033&0.906&0.936&\textcolor{myRed}{\textbf{0.898}}&\textcolor{myBlue}{0.035}\\
&&HIDANet&0.928&\textcolor{myRed}{\textbf{0.956}}&\textcolor{myRed}{\textbf{0.924}}&\textcolor{myBlue}{0.028}&\textcolor{myBlue}{0.910}&\textcolor{myRed}{\textbf{0.945}}&\textcolor{myBlue}{0.891}&\textcolor{myRed}{\textbf{0.034}}\\

\cmidrule{3-11} 
&&Ours & \textcolor{myRed}{\textbf{0.934}} & \textcolor{myBlue}{0.947} & \textcolor{myBlue}{0.913} & 0.029 & 0.908 & \textcolor{myBlue}{0.941} & 0.889 & 0.039 \\
\midrule
% \multicolumn{17}{l}{\textbf{Performance of RGB-D Models under Missing RGB}}\\
\multirow{6}[1]{*}{\emptycircle} &\multirow{6}[1]{*}{\solidcircle}
&D3Net&0.768&0.688&0.640&0.105&0.619&0.599&0.525&0.158\\
&&DIGR&0.646&0.598&0.486&0.170&0.585&0.537&0.374&0.190\\
&&C2DFNet&0.673&0.704&0.636&0.151&\textcolor{myBlue}{0.667}&0.649&\textcolor{myBlue}{0.620}&0.145\\
&&CIRNet&\textcolor{myBlue}{0.803}&0.768&0.733&0.109&0.642&0.617&0.514&0.158\\
&&SPSN&0.700&0.703&0.651&0.117&0.615&0.634&0.529&0.139\\
&&HIDANet&0.802&\textcolor{myBlue}{0.826}&\textcolor{myBlue}{0.769}&\textcolor{myBlue}{0.082}&0.663&\textcolor{myBlue}{0.732}&0.611&\textcolor{myBlue}{0.134}\\
\cmidrule{3-11} 
&&Ours&\textcolor{myRed}{\textbf{0.856}} & \textcolor{myRed}{\textbf{0.878}} & \textcolor{myRed}{\textbf{0.814}} & \textcolor{myRed}{\textbf{0.066}}&\textcolor{myRed}{\textbf{0.725}}&\textcolor{myRed}{\textbf{0.773}}&\textcolor{myRed}{\textbf{0.659}}&\textcolor{myRed}{\textbf{0.120}} \\
\midrule
\multirow{6}[1]{*}{\solidcircle} &\multirow{6}[1]{*}{\emptycircle}
% & SPNet & 0.857 & 0.910 & 0.864 & 0.063 & 0.913 & 0.953 & 0.900 & 0.027 & 0.920 & 0.968 & 0.909 & 0.021 & 0.898 & 0.938 & 0.903 & 0.045 & 0.903 & 0.943 & 0.899 & 0.040\\
& D3Net &0.883 & 0.904 & 0.847 & 0.054 & 0.892 & 0.915 & 0.855 & 0.051 \\
&&DIGR  & 0.850 & 0.861 & 0.821 & 0.063 & 0.876 & 0.894 & 0.855 & 0.052 \\
&&C2DFNet & 0.878 & 0.911 & 0.868 & \textcolor{myBlue}{0.048} & 0.890 & 0.922 & \textcolor{myBlue}{0.876} & \textcolor{myBlue}{0.041} \\
&&CIRNet & 0.892 & 0.900 & 0.872 & 0.056 & 0.891 & 0.901 & 0.869 & 0.055 \\
&&SPSN  & \textcolor{myBlue}{0.893} & 0.915 & 0.870 & \textcolor{myRed}{\textbf{0.043}} & 0.896 & 0.929 & \textcolor{myRed}{\textbf{0.879}} & \textcolor{myBlue}{0.041} \\
&&HIDANet  & \textcolor{myRed}{\textbf{0.894}} & \textcolor{myRed}{\textbf{0.929}} & \textcolor{myBlue}{0.886} & \textcolor{myBlue}{0.048} & \textcolor{myBlue}{0.897} & \textcolor{myBlue}{0.931} & \textcolor{myBlue}{0.876} & 0.042 \\
\cmidrule{3-11} 

&&Ours & \textcolor{myRed}{\textbf{0.894}} & \textcolor{myBlue}{\textbf{0.923}} & \textcolor{myRed}{\textbf{0.874}} &0.050&\textcolor{myRed}{\textbf{0.900}}&\textcolor{myRed}{\textbf{0.933}}&0.874&\textcolor{myRed}{\textbf{0.040}} \\
\midrule
\multicolumn{2}{c|}{\multirow{7}{*}{\textbf{Average}}}
& D3Net  &0.850 & 0.838 & 0.792 & 0.067 & 0.803 & 0.813 & 0.753 & 0.083 \\
&&DIGR  & 0.810 & 0.800 & 0.736 & 0.086 & 0.792 & 0.788 & 0.703 & 0.092 \\
&&C2DFNet & 0.819 & 0.850 & 0.801 & 0.079 & 0.820 & 0.836 & \textcolor{myBlue}{0.793} & 0.075 \\
&&CIRNet & 0.873 & 0.867 & 0.833 & 0.067 & 0.816 & 0.816 & 0.752 & 0.084 \\
&&SPSN  & 0.835 & 0.851 & 0.811 & 0.064 & 0.806 & 0.833 & 0.769 & 0.072 \\
&&HIDANet  & \textcolor{myBlue}{0.875} & \textcolor{myBlue}{0.904} & \textcolor{myBlue}{0.860} & \textcolor{myBlue}{0.053} & \textcolor{myBlue}{0.823} & \textcolor{myBlue}{0.869} & \textcolor{myBlue}{0.793} & \textcolor{myBlue}{0.070} \\

    \cmidrule{3-11} 
    &&Ours  & \textcolor{myRed}{\textbf{0.893}}  & \textcolor{myRed}{\textbf{0.916}}  & \textcolor{myRed}{\textbf{0.867}}  & \textcolor{myRed}{\textbf{0.048}}  & \textcolor{myRed}{\textbf{0.844}}  & \textcolor{myRed}{\textbf{0.882}}  & \textcolor{myRed}{\textbf{0.804}}  & \textcolor{myRed}{\textbf{0.066}}
\\
\midrule
\multicolumn{2}{c|}{\multirow{7}{*}{ \textbf{Average Drop}}}
& D3Net  & \textcolor{myBlue}{-0.075} & -0.125 & -0.147 & 0.039 & -0.144 & -0.168 & -0.189 & 0.064 \\
&&DIGR   & -0.185 & -0.212 & -0.247 & 0.091 & -0.184 & -0.219 & -0.266 & 0.086 \\
&&C2DFNet  & -0.132 & -0.129 & -0.146 & 0.061 & -0.125 & -0.152 & \textcolor{myBlue}{-0.136} & 0.055 \\
&&CIRNet & -0.078 & -0.099 & \textcolor{myBlue}{-0.093} & 0.048 & -0.148 & -0.171 & -0.183 & 0.069 \\
&&SPSN  & -0.116 & -0.127 & -0.152 & 0.047 & -0.151 & -0.155 & -0.194 & 0.055 \\
&&HIDANet & -0.080 & \textcolor{myBlue}{-0.078} & -0.097 & \textcolor{myBlue}{0.037} & \textcolor{myBlue}{-0.114} & \textcolor{myBlue}{-0.114} & -0.148 & \textcolor{myBlue}{0.054} \\

    \cmidrule{3-11} 
    &&Ours & \textcolor{myRed}{\textbf{-0.054}}  & \textcolor{myRed}{\textbf{-0.047}}  & \textcolor{myRed}{\textbf{-0.069}}  & \textcolor{myRed}{\textbf{0.029}}  & \textcolor{myRed}{\textbf{-0.096}}  & \textcolor{myRed}{\textbf{-0.088}}  & \textcolor{myRed}{\textbf{-0.113}}  & \textcolor{myRed}{\textbf{0.041}}
 \\

\bottomrule
\end{tabular}
\label{supp_rgb-d-2}
\end{table*}
\subsection{Qualitative Evaluation}
\label{sec:Qualitative Evaluation}
\cref{supp_visual_depth} provides a visualization of RGB-D. When all modalities are available, our model demonstrates good performance, even in images with poor depth quality. In the absence of depth or RGB, our model outperforms other models in extracting valuable information from the remaining modalities. Particularly, when RGB is missing, our model can still extract significant objects from the depth image, which other comparative methods cannot achieve.
\section{Limitations and Future Works}
\label{sec:limitation}
% Despite our model's success in handling noisy inputs and missing-modalitiy. Our model still has some limitations, as seen in the example in \cref{supp_failure}. We have identified two aspects where our model needs further improvement. Firstly, there are some deficiencies in semantic recognition methods, as seen in (a) and (b) of \cref{supp_failure}, where the model fails to correctly identify prominent objects in the images. Secondly, our model lacks some methods for handling intricate details, such as edge supervision, leading to instances like (c) and (d) in \cref{supp_failure}, where certain object edge details cannot be detected effectively.
Our model has effectively addressed issues of low-quality image inputs and modality-missing in the dual-modal salient object detection domain. In practice, these problems persist in other communities such as multi-modal instance segmentation and semantic segmentation. In the future, we will continue to explore the issue of modality-missing in other multi-modal domains.

% \bibliography{main}
\end{document}